\documentclass[runningheads]{llncs}

\usepackage{accv}

\usepackage{accvabbrv}

\usepackage{graphicx}
\usepackage{booktabs}

\usepackage{times}
\usepackage{epsfig}
\usepackage{amsmath}
\usepackage{multirow}
\usepackage{array}

\newcommand{\sftype}[1]{{\textsf{\small #1}}}
\usepackage{pifont}

\newcommand{\xmark}{\ding{55}}

\usepackage[accsupp]{axessibility}  % Improves PDF readability for those with disabilities.

\usepackage[pagebackref,breaklinks,colorlinks,citecolor=accvblue]{hyperref}
\usepackage{hyperref}

\usepackage{orcidlink}

\begin{document}

\title{Improving Text-based Person Search via Part-level Cross-modal Correspondence} 


\author{Jicheol Park \orcidlink{0009-0004-7899-6802} \and
Boseung Jeong \orcidlink{0000-0001-9382-3396} \and
    Dongwon Kim \orcidlink{0000-0003-1147-5274} \and
    Suha Kwak \orcidlink{0000-0002-4567-9091}}

\authorrunning{J.~Park et al.}

\institute{ Pohang University of Science and Technology (POSTECH), South Korea\\
\email{\{jicheol, boseung01, kdwon, suha.kwak\}@posetech.ac.kr
}}

\maketitle
%[Abstract]
%Text-based person search task 설명
%Main challenge
%1) modality gap
%2) fine-grained recognition

%Text-based person search is the

\begin{abstract}
Text-based person search is the task of finding person images that are the most relevant to the natural language text description given as query.
The main challenge of this task is a large gap between the target images and text queries, which makes it difficult to establish correspondence and distinguish subtle differences across people.
To address this challenge, we introduce an efficient encoder-decoder model that extracts coarse-to-fine embedding vectors which are semantically aligned across the two modalities without supervision for the alignment.
There is another challenge of learning to capture fine-grained information with only person IDs as supervision, where similar body parts of different individuals are considered different due to the lack of part-level supervision.
To tackle this, we propose a novel ranking loss, dubbed commonality-based margin ranking loss, which quantifies the degree of commonality of each body part and reflects it during the learning of fine-grained body part details.
As a consequence, it enables our method to achieve the best records on three public benchmarks.
\end{abstract}
% person search and text-based person search
% difference between text-based person search and image or attribute-based person search
% challenges in text-based person search
% 1) modality gap, fine-grained matching
% 2) ID level matching in local feature

%Motivation
%Text-based person search requires local information to capture subtle differences between persons.
%However, extracting and aligning local features across two modalities is challenging without local supervisions.

%Contribution
%Extracting semantically aligned embeddings using a shared token and decoder across two modalities. (coarse-embeddings)
%Learning semantically aligned embeddings with commonality-based margin ranking loss. (fine-embeddings)

\section{Introduction}

Person search is the task of finding individuals within a vast collection of images based on queries describing visual characteristics in images~\cite{Sun_2018_ECCV, Kalayeh_2018_CVPR, Ali_2018_ECCV, Zhong_2018_CVPR, Wang_2018_ECCV}, attribute sets~\cite{yin2017adversarial, dong2019person, cao2020symbiotic, jeong2021asmr}, and natural language~\cite{li2017person, li2017identity, zhang2018deep, gao2021contextual, ding2021semantically, suo2022simple}.
It has played vital roles in public safety applications, such as identifying criminals in videos and finding missing people using multiple surveillance cameras that have non-overlapping fields of view.

Text-based person search refers to the person search task that employs free-form text as query. %to retrieve individuals within a large collection of images.
This task enables efficient and effective person search thanks to the flexible and user-friendly query acquisition.
Also, text queries are often more suitable for person search in the wild than other types of queries such as images~\cite{Sun_2018_ECCV, Kalayeh_2018_CVPR} and attributes~\cite{yin2017adversarial, jeong2021asmr}: Image queries are not accessible when eyewitness memory is the only evidence for identification, and the expression power of attribute-based queries is restricted strictly by a predefined set of attributes.
However, the advantages of text queries come with an additional challenge, the large gap between image and text modalities.
Due to the modality gap, it becomes even more difficult to extract semantically corresponding information necessary for distinguishing subtle differences across people from both modalities.
% Due to this modality gap, it becomes even more difficult to establish correspondences between target and query, 
% and consequently to distinguish subtle differences across people.

% To overcome the modality gap, prior studies have employed joint image-text embedding 
% % that aims at representing 
% to represent
% data of both modalities in a common embedding space~\cite{frome2013devise, kiros2014unifying, faghri2017vse++}.
% They have emphasized, in particular, the development of loss functions to mitigate the modality gap.
% However, they have not paid sufficient attention to the model architecture design for reducing the gap during the encoding process of each modality, and thus leave room for further improvement.

\begin{figure*}[t!]
    \vspace{1mm}
    \centering
    \includegraphics[width=\textwidth]{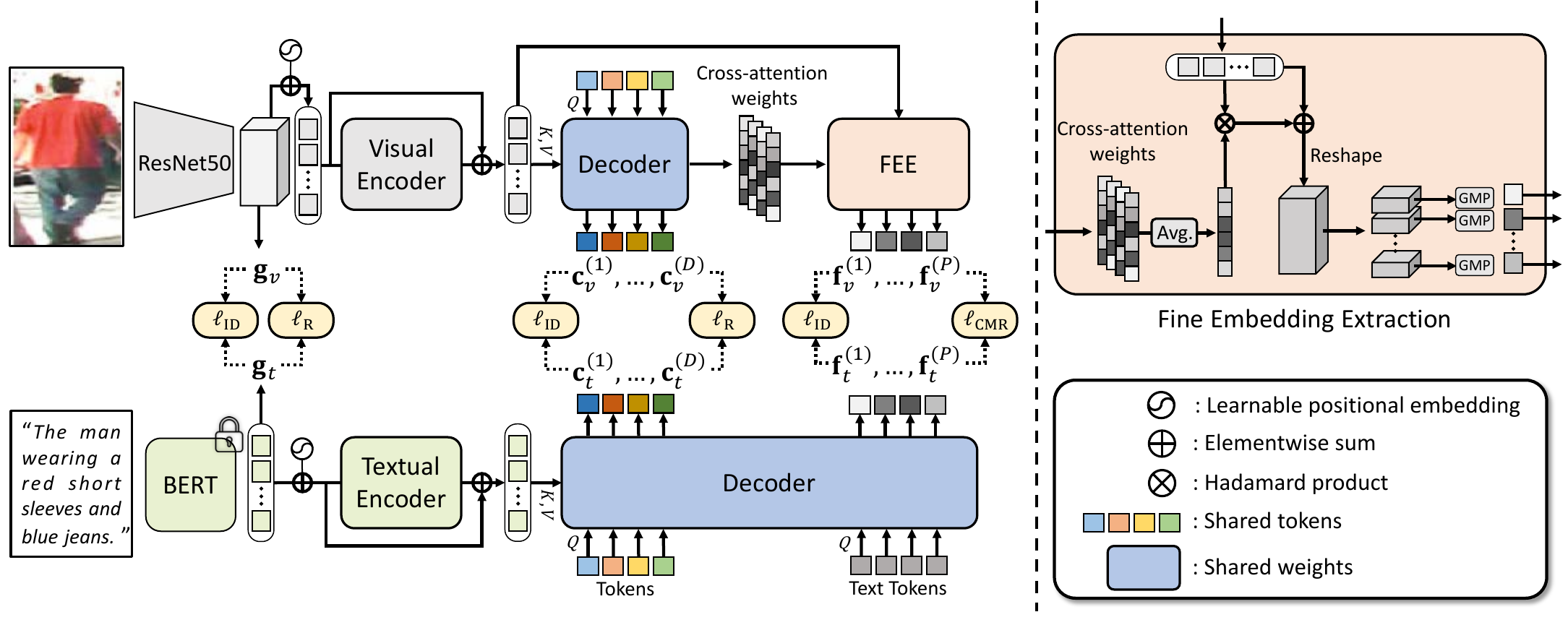}
    \caption{
    %\bstodo{Temporary figure. Must be revised. Loss and Slot attention module will be revised and added.}
    The overall pipeline of our method. 
    Given an image and a text description of a person as input, the corresponding backbone models (ResNet50 and BERT) extract visual and textual features.
    Global embeddings of the image and the text description ($\mathbf{g}_v$ and $\mathbf{g}_t$) are obtained by global max pooling over the visual and textual features. 
    Coarse embeddings of the image and text modalities ($\mathbf{c}_v$ and $\mathbf{c}_t$) are produced by an encoder-decoder taking the visual and textual features, and a set of learnable tokens as input. 
    Fine embeddings of the image ($\mathbf{f}_v$) are obtained by horizontally dividing the feature map. 
    % produced by applying Hadamard product between averaged attention weights from the decoder and output of visual encoder followed by a residual connection. 
    Meanwhile, the fine embeddings of the text description ($\mathbf{f}_t$) are extracted by the decoder with another set of tokens, namely text tokens. 
    The global and coarse embeddings of both image and text description are aligned by conventional identity classification loss ($\ell_\textrm{ID}$) and ranking loss ($\ell_\textrm{R}$). 
    On the other hand, the fine embeddings are aligned by identity classification loss and commonality-based margin ranking loss ($\ell_\textrm{CMR}$).
    }
    \label{fig:architecture}
\end{figure*}

To overcome above issue, we introduce an efficient encoder-decoder model based on multi-head attention~\cite{vaswani2017attention};
its overall architecture is illustrated in Fig.~\ref{fig:architecture}.
% , and a new loss function for leaning joint image-text embedding at the local level.
The key aspect of our model is that it extracts coarse-to-fine embedding vectors which are semantically aligned between the two modalities without supervision for the alignment but by its architecture dedicated to that purpose.
%To address this challenge, we introduce an efficient encoder-decoder model that extracts coarse-to-fine embedding vectors which are semantically aligned across the two modalities without supervision for the alignment.
To be specific, although features of the two modalities are computed by separate encoders, they are aggregated into a fixed number of embedding vectors by \emph{tokens common to both modalities} in the shared decoder.
By matching such embeddings of the same token during training, each token is learned to capture the same entity from inputs of the two different modalities and thus establishes correspondence between them. 
We call these embeddings aligned by the shared tokens \emph{coarse embeddings} since they are attained by the vanilla cross-attention mechanism that holistically investigates input features to draw attention.

In addition to the coarse embeddings, our model extracts \emph{fine embeddings}, which capture details of body parts, to distinguish subtle differences between people.
Learning to extract fine embeddings is even more challenging since the model should be aware of body parts with no explicit supervision.
To overcome this difficulty in the image modality, we divide the feature map of input image into multiple cells with no overlap in the vertical direction and consider each cell as a body part; each fine embedding of the image modality is then extracted from each cell.
This heuristic, however, is not viable in the text modality due to the varying structures of texts.
We thus employ additional \emph{text tokens} that extract fine embeddings for text query through cross-attention.
Each text token is learned to capture specific part information by matching its fine embedding with the associated visual fine embedding during training.

% Although the fine embeddings are necessary for complementing person representation,
% complementary to their coarse counterparts, 
%However, given person IDs as the only supervision, learning the fine embeddings is not straightforward since similar body parts of different people are considered different due to the lack of part-level supervision (see Fig.~\ref{fig:Loss}).
\iffalse
However, learning the fine embeddings with only person IDs as the supervision is challenging due to the presence of common body parts shared between individuals, which cannot be distinguished based solely on person ID (see Fig.~\ref{fig:Loss}).
% body parts semantically close to each other are considered different when they come from different people (\ie, the same body parts of different identities are considered different.
We address this issue that hinders learning the fine embeddings by a new loss function, dubbed commonality-based margin ranking (CMR) loss.
The loss quantifies the degree of commonality of each fine embedding and reflects such degree during training. 
Specifically, it is adjusted on-the-fly by the degree of commonality so that fine embeddings with similar semantics are allowed to be close to each other in the embedding space although they are of different IDs. 
\fi
Learning the fine embeddings with only person IDs as the supervision is challenging due to the presence of common body parts shared between individuals, which cannot be distinguished based solely on person ID (see Fig.~\ref{fig:Loss}).
To distinguish semantically identical body parts and incorporate this information during training, we introduce a novel loss function called the commonality-based margin ranking (CMR) loss.
This loss quantifies the commonality for each fine embedding, indicating how frequently the corresponding body part is shared among individuals.
% Subsequently, the margin in the loss is adjusted on-the-fly based on the commonality.
% %margin풀어서 설명, part-level embedding learning에 push 강도를 조절, 기존에는 정해진 margin만큼 push 했으나
% This dynamic adjustment enables fine embeddings representing commonly shared body parts to be close to each other, even though they are assigned with different person IDs.
Subsequently, the margin in the loss is adjusted on-the-fly based on the commonality, which enables fine embeddings representing commonly shared body parts to be close to each other, even though they are assigned with different person IDs.

%실제 실험 세팅인 unseen setting에서 generalization 측면으로...    
% \fi

Our method was evaluated on three public benchmarks~\cite{li2017person, ding2021semantically, zhu2021dssl}, where it clearly outperforms all existing methods thanks to the rich representation based on coarse and fine embeddings.
The main contribution of our work is three-fold:
\begin{itemize}
   \item We propose an efficient architecture for text-based person search that effectively closes the modality gap by producing coarse-to-fine multiple person embeddings which are semantically aligned between the two modalities.
   \item We also introduce a new loss function that allows our model to learn to capture fine-grained information (\ie, fine embeddings) appropriately using only person IDs as supervision.
   \item Our method achieved the state of the art on the three public benchmarks.
\end{itemize}

\section{Related Work}
\subsection{Image-Text Matching}
%\subsection{Image-Text Matching}
The task of Image-Text matching, which involves retrieving relevant images based on textual descriptions and vice versa, is a crucial problem in the intersection of vision and language. 
Early studies~\cite{frome2013devise,kiros2014unifying, faghri2017vse++} focused on learning a visual semantic embedding (VSE) to represent both visual and text modalities in a joint embedding space that learned from hinge-based triplet ranking loss. %Frome, Kiros, Faghri
%Faghri
~\cite{faghri2017vse++} utilized online hard negative mining in the triplet ranking loss and improved discriminability of the embedding space.
Moreover, recent studies~\cite{lee2018stacked, li2019visual, chen2020imram, diao2021similarity, chen2021learning, han_2021_SIGIR} have dedicated to exploiting local information (\eg, image regions and text words) to utilize more fine-grained exploration, and have achieved great improvements.
Some studies propose a new hashing method according to the modality for more efficient image-text retrieval~\cite{daph_2023_sigir} or deal with the data imbalance problem between modalities for robust image-text matching~\cite{pan_2021_SIGIR}. 
%However, these studies have yet to adequately address the specific characteristics inherent in text-based person search tasks, such as the extremely low inter-class variance and the structure of human body parts.

\subsection{Text-Based Person Search}
% Global feature for image modality (GNA-RNN, TextReID)
%   (-) Hard to caputre fine-grained information which is crucial to the text-based person search task.
%   (-) Suffer from the background clutter.
In recent years, the task of text-based person search has gained significant attention in the computer vision community. 
Li~\etal,~\cite{li2017person} proposed a gated neural attention-based recurrent neural network (GNA-RNN) for learning the affinity between text descriptions and images in the person search task. 
In addition, they provided a benchmark dataset CUHK-PEDES for the evaluation of the proposed model.
Zhang~\etal,~\cite{zhang2018deep} proposed cross-modal projection matching loss (CMPM) and cross-modal projection classification (CMPC) loss, for learning deep discriminative image-text embeddings.
% Han~\etal~\cite{han2021textreid} proposed a method that learns more discriminative features by using the proposed cross-modal momentum contrastive learning strategy. 
While these methods mitigate the modality gap between visual and text modalities by learning discriminative embeddings, these methods primarily focus on global representations of person images and are thus not capable of capturing distinctive local details, leading to limited performance in the text-based person search task.

To address the above problem, a line of studies focuses on mining fine-grained representations.
One of the prominent examples of exploiting fine-grained information is to cut human images horizontally and use them as local features~\cite{gao2021contextual, niu2020improving, wang2021text, chen2022tipcb, ding2021semantically, suo2022simple, beat}.
Specifically, Gao~\etal,~\cite{gao2021contextual} conducted joint alignment over multi-scale representation via the contextual non-local attention mechanism.
Wang~\etal,~\cite{wang2021text} proposed a multi-granularity embedding learning model that exploits the representations of human body parts in different granularity.
Chen~\etal,~\cite{chen2022tipcb} built a new dual-path local alignment network to learn visual and textual local representations, which is optimized by a multi-stage cross-modal matching strategy.
Ding~\etal,~\cite{ding2021semantically} utilized the attention mechanism to capture the relationships between body parts and to select words in sentences.
Despite the efforts to make informative local features, they still remain the risk of including background clutter and the problem 
that extracting local features from text description such as the horizontal segmentation is not available.
% of extracting heuristic local features from text modality due to their diverse nature.

Several studies tried to utilize useful information (\eg, human attributes and human keypoints) via external tools~\cite{aggarwal2020text, wang2020vitaa, jing2020pose}.
Aggarwal~\etal~\cite{aggarwal2020text}, introduced an auxiliary task, attribute recognition, to bridge the modality gap between the image-text inputs, and to improve the representation learning.
Wang~\etal,~\cite{wang2020vitaa} introduced an auxiliary attribute segmentation to align the visual features with the textual attributes parsed from the sentences.
Jing~\etal,~\cite{jing2020pose} proposed a new multi-granularity attention network to learn the latent semantic alignment between local visual and textual information with human pose estimation. 
However, these approaches have inevitable limitations of high computational cost and dependence on the performance of external tools for local feature extraction.
To avoid these limitations, Suo~\etal,~\cite{suo2022simple} proposed the simple and robust correlation filtering (SRCF) method to extract crucial local features and adaptively align them without any external tools. 
Specifically, it produces the local features via denoising filters and dictionary filters. 
Similar to our coarse embeddings, Shao~\etal,~\cite{lgur} proposed the unified representations with multi-head attentions.
However, obtaining their unified representations involves the learning of costly reconstruction tasks.
Moreover, it does not address the fine embeddings, which play a crucial role in distinguishing subtle differences to find the target person in the extensive search space.

Although these approaches shed light on the importance of exploiting local features for each modality, there is still a large room for further improvement.
They suffer from aligning the local features of the two modalities due to the hidden correspondences between image and text description. 
% Aligning the local features of the two modalities still remains challenging due to hidden correspondences between image and text descriptions.
Moreover, these methods neglect considering a characteristic of local features that can be shared across multiple person identities; it makes learning a discriminative embedding space more difficult.
% ,learning local feature alignment using identity labels is also challenging due to the above approaches have less focused on the loss functions.
Our method can overcome these problems thanks to a new encoder-decoder architecture that extracts semantically aligned embedding vectors from the two modalities without supervision for the alignment and a novel loss function that quantifies the degree of commonality of fine embedding and reflects it through margins.

% Local feature for image modality
%   -external tool (ViTAA, PMA, Lapscore
%       (-) Require an extra network, module, or data.
%       (-) Inevitable error accumulation (in other words, the depend on the reliability of external tools). 
%   -heuristic feature (SRCF, TIPCB, MGEL
%       (-) Still suffer from background clutter.
%       (-) Inconsistency of human part's location (?): Human body part does not always appear in the same divided part.
%       (-) The contents of the part features can be shared between different people.

% ------------------------------------------------------------------------------------------------------------- %
% Notations. 
% Feature map or Sequential features: CAPITAL LETTER
% Feature (embedding) vector: small letter
% Modality indicator: Superscript
% Number of somethings: subscript
% Feature map from the backbone: F
% Output of the encoder: $\tilde{F}$
% Output of the decoder: L (?)
% Output of modality fine feature extractor: P=\{p_i\}_{i=1}^{|P|}
% ------------------------------------------------------------------------------------------------------------- %
\section{Proposed Method}
\iffalse
In this section, we first describe the details of the model architecture which is composed of image and text backbones, two encoders for image and text modalities, and a decoder shared across the two modalities as illustrated in Fig.~\ref{fig:architecture}.
Then, we present the fine embedding extraction. 
Finally, the proposed loss function is described in detail.
We provide an overview of the notation used in this section in Table~\ref{tab:notation table}.
\fi
This section describes the details of the model architecture (section~\ref{sec:architecture}), the fine embedding extraction (section~\ref{sec:fine_embedding_extraction}), the proposed loss function (section~\ref{sec:loss}), and the inference process (section~\ref{sec:inference}) of our method.
% Table~\ref{tab:notation table} provides an overview of the notation used in this section.

% \input{tables/tab_notation}

\subsection{Model Architecture}
\label{sec:architecture}
\noindent\textbf{Image and Text Backbones.}
Following conventional strategies~\cite{gao2021contextual, wu2021lapscore, suo2022simple, saf, ivt}, we adopt ResNet-50~\cite{resnet} or ViT-B/16~\cite{dosovitskiy2020image} as the backbone for the image modality and the uncased BERT~\cite{devlin2018bert} model as the backbone for the text modality. 
In detail, given an image \textit{I} and a text description \textit{T} as input, the corresponding backbones extract visual and textual features $\mathbf{V}\in \mathbb{R}^{w\times h\times d}$ and $\mathbf{T}\in \mathbb{R}^{l\times d}$, where the $w\times h$ indicates the spatial resolution of the visual feature map, and the maximum number of words in the text description and dimension of their features are denoted by $l$ and $d$, respectively.
The global embeddings of image and text modalities are obtained by global max pooling (GMP) over $\mathbf{V}$ and $\mathbf{T}$, and denoted by $\mathbf{g}_v$ and $\mathbf{g}_t$.

\noindent\textbf{Visual and Textual Encoders.}
%In image modality, after the image backbone, 
We flatten the visual feature map $\mathbf{V}$ into $\bar{\mathbf{V}} \in \mathbb{R}^{r\times d}$, where $r=wh$.
Then, we inject learnable position embeddings $\mathbf{E}^{pos}_{v} \in \mathbb{R}^{r\times d}$ into $\bar{\mathbf{V}}$ by $\mathbf{M}_v = \bar{\mathbf{V}} + \mathbf{E}^{pos}_{v}$.
%flatten and positional embedding 관련 간소화 가능
% In text modality, similar to image modality, we obtain the positional encoded textual features $\mathbf{M}_t$ by adding learnable positional encoding $\mathbf{E}^{pos}_{t} \in \mathbb{R}^{l\times d}$ to $\mathbf{T}$, $\mathbf{M}_t = \mathbf{T} + \mathbf{E}^{pos}_{t}$. 
% Then the positional encoded modality features $\mathbf{M}_v$ and $\mathbf{M}_t$ are fed into each modality encoder.
The position-encoded visual features $\mathbf{M}_v$ are fed into the visual encoder.
The visual encoder employs a simple multi-head self-attention (MHSA) structure~\cite{vaswani2017attention}, composed of $n$-head Self-Attention (SA) operations. For $i$-th head, $\textrm{SA}_i$ operation is conducted by query $\mathbf{Q}_i \in \mathbb{R}^{r\times d_h} $, key $\mathbf{K}_i \in \mathbb{R}^{r\times  d_h}$, and value $\mathbf{V}_i \in \mathbb{R}^{r\times  d_h}$ obtained by linear projections of visual features $\mathbf{M}_v \in \mathbb{R}^{r\times d}$:
%\begin{equation}
\begin{align} 
    &\textrm{SA}_i(\mathbf{M}_v) = A_i\mathbf{V}_i, \\
    &\textrm{where}~ A_i = \textrm{softmax}\Big(\mathbf{Q}_i\mathbf{K}_i^{\top}/\sqrt{d_h}\Big), \nonumber\\
    &\mathbf{Q}_i = \mathbf{M}_v\mathbf{W}_{i}^{Q}, \quad \mathbf{K}_i = \mathbf{M}_v\mathbf{W}_{i}^{K}, \quad \mathbf{V}_i = \mathbf{M}_v\mathbf{W}_{i}^{V}, \nonumber 
    \label{eq:self-attention}
\end{align} 
and $\mathbf{W}_{i}^{Q} \in \mathbb{R}^{d\times d_h}$, $\mathbf{W}_{i}^{K} \in \mathbb{R}^{d\times d_h}$, and $\mathbf{W}_{i}^{V} \in \mathbb{R}^{d\times d_h}$ are the linear projection weights for query, key, and value, respectively. 
$A_i \in \mathbb{R}^{r\times r}$ is dot-product attention scaled with $\sqrt{d_h}$, where $d_h$ is set to $d/n$ following~\cite{vaswani2017attention}. 
Then MHSA is calculated by:
\begin{equation}
    \textrm{MHSA}(\mathbf{M}_v) = [\textrm{SA}_1(\mathbf{M}_v), \cdots ,\textrm{SA}_n(\mathbf{M}_v)]\mathbf{W}^{O}, 
    % \textrm{MHSA}(\mathbf{M}_v) = [\textrm{SA}_1(\mathbf{M}_v), \textrm{SA}_2(\mathbf{M}_v), \cdots ,\textrm{SA}_n(\mathbf{M}_v)]\mathbf{W}^{O}, 
\label{eq:multi-head self-attention}
\end{equation} 
where $[\cdot,\cdot]$ denotes the concatenation operation, and $\mathbf{W}^{O} \in \mathbb{R}^{nd_h\times d}$ is a linear projection for the multiple heads.
% Each visual and textual encoder comprises a single MHSA and a residual connection. 
% These encoders take modality features $\mathbf{F}_v$ and $\mathbf{F}_t$ as input and extract the self-attended modality features $\tilde{\mathbf{F}}_v$ and $\tilde{\mathbf{F}}_t$, respectively:
The visual encoder comprises a single MHSA and a residual connection that takes $\mathbf{M}_v$ as input, and extracts the self-attended visual features, $\tilde{\mathbf{M}}_v$, as follows:  
\begin{eqnarray} 
    \tilde{\mathbf{M}}_v = \textrm{Encoder}_v(\mathbf{M}_v)  = \mathbf{M}_v + \textrm{MHSA}_v(\mathbf{M}_v).
\label{eq:visual and textual encoder}
\end{eqnarray}
Likewise, given a textual feature sequence by BERT, the textual encoder extracts the self-attended textual features, $\tilde{\mathbf{M}}_t\in\mathbb{R}^{l \times d}$, where $l$ denotes the number of words in the text description.

\vspace{1mm}\noindent\textbf{Modality-sharing Decoder.}
Similar to the encoders, the decoder employs a multi-head attention (MHA) structure with $n$-head attention operations.
However, unlike the modality-specific encoders, the decoder computes attention using a set of learnable tokens 
$\mathbf{D} =\{\mathbf{d}^{(i)} \}^D_{i=1}$ as query, where $D$ denotes the number of the tokens, and extracts corresponding coarse embeddings from each modality.
% $\mathbf{D}\in\mathbb{R}^{D\times d}$ where $D$ denotes the size of the token embedding set, and extracts corresponding local features from each modality.
The parameters of the decoder and the tokens are shared across the modalities.
Our decoder takes the set of tokens 
$\mathbf{D}$ and the self-attended features $\tilde{\mathbf{M}}$ as input, and compute cross-attention between them to extract the coarse embeddings $\mathbf{C}=\{\mathbf{c}^{(i)}\}_{i=1}^{D}$ for each modality as follows:
% \begin{eqnarray}
%     &\textrm{A}_i(\mathbf{D},\mathbf{F}) = A_i\mathbf{V}_i, \quad A_i = \textrm{softmax}(\mathbf{Q}_i\mathbf{K}_i^{\top}/\sqrt{d_h}),\nonumber\\
%     &\mathbf{Q}_i = \mathbf{D}\mathbf{W}_{i}^{Q}, \quad \mathbf{K}_i = \tilde{\mathbf{F}}\mathbf{W}_{i}^{K}, \quad \mathbf{V}_i = \tilde{\mathbf{F}}\mathbf{W}_{i}^{V},
% \label{eq:decoder attention}    
% \end{eqnarray}
% \begin{eqnarray}
%     \textrm{MHA}(\mathbf{D}, \tilde{\mathbf{F}}_v) = [\textrm{A}_1(\mathbf{D}_1,\mathbf{F}_v), \cdots ,\textrm{A}_n(\mathbf{D}_n,\mathbf{F}_v)]\mathbf{W}^{O},
% \end{eqnarray}
\begin{align}
    &\mathbf{C}_v = \textrm{Decoder}(\mathbf{D},\tilde{\mathbf{M}}_v)  = \textrm{MHA}(\mathbf{D}, \tilde{\mathbf{M}}_v), \nonumber \\
    &\mathbf{C}_t = \textrm{Decoder}(\mathbf{D},\tilde{\mathbf{M}}_t)  = \textrm{MHA}(\mathbf{D}, \tilde{\mathbf{M}}_t).
    \label{eq:modality-agnostic decoder}
\end{align}
The proposed decoder enables the extraction of coarse embeddings that ensure correspondence between the modalities for cross-modal semantic alignment without dedicated supervision.
This is achieved by sharing the learnable tokens and the decoder parameters across the two modalities.
%The details of our encoder-decoder architecture are illustrated in Fig.~\ref{fig:encoder_decoder}.

\subsection{Fine Embedding Extraction}
\label{sec:fine_embedding_extraction}
For fine-grained recognition, we extract foreground visual features  $\ddot{\mathbf{M}}_v$, eliminating extraneous background information.
%To do this, we obtain foreground attention map $A_\textrm{avg}$, by averaging over cross-attention maps between the shared tokens $\mathbf{D}$ and self-attended visual features $\tilde{\mathbf{M}}_v$ from the decoder in Eq. ~\eqref{eq:modality-agnostic decoder}.
To do this, we obtain foreground attention map $A_\textrm{avg}$, by averaging cross-attention weights obtained while extracting coarse embeddings of image modality in Fig.~\ref{fig:architecture}.
More specifically, the cross-attention weights are computed by shared tokens $\mathbf{D}$ as query and self-attended visual features $\tilde{\mathbf{M}}_v$ as key from the decoder in Eq. ~\eqref{eq:modality-agnostic decoder}.
%The $A_\textrm{avg}$ highlights the comprehensive human area by averaging attention maps obtained while extracting coarse embeddings.
%The $A_\textrm{avg}$ highlights the comprehensive human area by averaging attention maps obtained while extracting coarse embeddings; empirical results in Fig.~\ref{fig:attention}(a) visually prove this observation.
With the Hadamard product between the average of cross-attention weights $A_\textrm{avg}$ and the self-attended visual feature $\tilde{\mathbf{M}}_v$, we obtain foreground visual features $\ddot{\mathbf{M}}_v$:
% computed by Eq.~\eqref{eq:decoder attention}:
\begin{eqnarray}
    \ddot{\mathbf{M}}_v = \tilde{\mathbf{M}}_v + A_\textrm{avg} \otimes \tilde{\mathbf{M}}_v,
    \label{spatial attended visual features}
\end{eqnarray}
where $\otimes$ is Hadamard product.
Exploiting the $A_\textrm{avg}$ allows to ignore some background clutters, it is empirically demonstrated in Fig~\ref{fig:attention}(a).
% For more fine-grained recognition, we extract specific fine embeddings from input image. Initially, we obtain person attended visual features $\hat{\mathbf{M}}_v$ using the Hadamard product between the self-attended visual feature $\tilde{\mathbf{M}}_v$ and the average cross-attention $A_\textrm{avg}$.  of
% $A_\textrm{avg}$ is the average across columns of $A \in\mathbb{R}^{d\times r}$, the cross-attention score matrix between $\mathbf{D}$ and $\tilde{\mathbf{M}}_v$, and is given by $A_\textrm{avg} = \Sigma^d_{i=1}\mathbf{a}_i / d$, where $\mathbf{a}_i$ is $i$-th column vector of $A$.
% obtained between the tokens $\mathbf{D}$ and the self-attended visual features $\tilde{\mathbf{M}}_v$:
% % computed by Eq.~\eqref{eq:decoder attention}:
% \begin{eqnarray}
%     \hat{\mathbf{M}}_v = \tilde{\mathbf{M}}_v + A_\textrm{avg} \otimes \tilde{\mathbf{M}}_v,
%     \label{spatial attended visual features}
% \end{eqnarray}
% where $\otimes$ is Hadamard product.
Then $\ddot{\mathbf{M}}_v$ is uniformly divided into $P$ horizontal segments without overlapping denoted as $\ddot{\mathbf{M}}^{(i)}_v \in \mathbb{R}^{r'\times d}$, where $i\in \{1,\cdots,P\}$ and $r'=r/P$.
This horizontal division method to obtain part feature is the conventions of previous work~\cite{gao2021contextual, niu2020improving, wang2021text, chen2022tipcb, ding2021semantically, suo2022simple}.
Unlike previous work, we emphasize foreground features before division with foreground attention map $A_\textrm{avg}$ to ignore background clutter.
% Finally, we obtain part local features by conducting Global Max Pooling (GMP) over all the $\mathbf{M}^p_v$:
Finally, the visual fine embeddings, denoted by $\mathbf{F}_v = \{\mathbf{f}_v^{(i)} \}^P_{i=1}$, are obtained by applying global max pooling (GMP) to each $\ddot{\mathbf{M}}^{(i)}_v$.
The comprehensive process of fine embedding extraction of image modality is illustrated in Fig.~\ref{fig:architecture}(FEE).
% \begin{eqnarray}
% &\mathbf{p}^{(i)}_v = \textrm{GMP}(\mathbf{F}^{(i)}_v). 
% &  
% \label{visual fine features}
% \end{eqnarray}
% Unlike traditional approaches, our method employs the average attention obtained from the decoder when extracting coarse embeddings.
% This approach enables us to address vulnerability to the background clutter by attending to the comprehensive region of the person.

In the text modality, since extracting fine embeddings such as the horizontal segmentation is not accessible, the additional set of text tokens with $P$ size, $\mathbf{D}_t =\{\mathbf{t}_t^{(i)} \}^P_{i=1}$ are arranged to handle specific fine embeddings of the text description. 
% due to the difficulty of extracting relevant fine features, the additional $P$ text query dictionaries $\mathbf{D}_t \in\mathbb{R}^{P \times d}$ are arranged to handle specific fine features of the text modality. 
To be specific, the textual fine embeddings $\mathbf{F}_t = \{\mathbf{f}_t^{(i)} \}^P_{i=1}$  are extracted by the decoder with $\mathbf{D}_t$ and $\mathbf{\tilde{M}}_t$ as follows:
\begin{eqnarray}
\mathbf{F}_t = \textrm{Decoder}(\mathbf{D}_t,\mathbf{\tilde{M}}_t). 
% \mathbf{p}^{(i)}_t = \textrm{Decoder}(\mathbf{D}_t,\mathbf{\tilde{F}}_t). 
\label{textual fine features}
\end{eqnarray}
% and these extracted textual fine embeddings are aligned to corresponding visual fine embeddings of each other
% The extracted textual fine embedding $\mathbf{f}_t^{(i)}$ corresponds to specific visual fine embedding $\mathbf{f}_v^{(i)}$ for cross-modal semantic alignment.

The extracted textual fine embedding $\mathbf{f}_t^{(i)}$ from \textit{i}-th text token and the \textit{i}-th visual fine embedding $\mathbf{f}_v^{(i)}$ are to be aligned in the joint embedding space.
%Additionally, following the convention of previous work~\cite{ding2021semantically,suo2022simple}, we also adopt a non-local module for obtaining the non-local embeddings $\mathbf{n}_m$ by attention-based aggregating between the coarse and fine embeddings (\eg, $\mathbf{C}_m$ and $\mathbf{F}_m$), where $m$ is modality indicator.
Finally, a set of embeddings per modality for training and inference is denoted by $\mathcal{S}_m = \{\mathbf{g}_m, \mathbf{C}_m, \mathbf{F}_m \}$, where ${m\in\{v,t\}}$.
% our granularity of modality features for training and inference is denoted by $\mathcal{G} \in  \{\mathbf{g}_m, \mathbf{C}_m, \mathbf{F}_m, \mathbf{n}_m \}$, where ${m\in\{v,t\}}$.
%non local 수정

% \input{figures/fig_loss}

\subsection{Learning Objective}
\label{sec:loss}
Our model employs two primary types of loss functions.
The first type is identity classification loss $\mathcal{L}_\textrm{ID}$, which is based on prior research~\cite{niu2020improving, wang2021text,suo2022simple}:
\begin{eqnarray}
\ell_\textrm{ID}(\mathbf{e}) = -\boldsymbol{y} \log(\boldsymbol{p}), \quad \boldsymbol{p} = \textrm{softmax}(\mathbf{e}\mathbf{W}_\textrm{ID}),
\label{eq:id loss}
\end{eqnarray}
where $\boldsymbol{y}\in \mathbb{R}^c$ is the identity ground truth represented by a one-hot vector, $\mathbf{W}_\textrm{ID} \in \mathbb{R}^{d\times c}$ is a classifier shared by the corresponding embeddings $\mathbf{e} \in \mathcal{S}_m$ between the two modalities, $\boldsymbol{p} \in \mathbb{R}^c$ is identity classification scores, and $c$ is the number of identities.
The identity classification loss $\mathcal{L}_\textrm{ID}$ plays a role in classifying identity differences within each modality and sharing the classifier between the two modalities to ensure that the feature embeddings of the modalities become similar. 
This loss applies to the set of embeddings per modality $\mathcal{S}_m$:
\begin{equation}
\mathcal{L}_\textrm{ID}(\mathcal{S}_m)={\ell}_\textrm{ID}(\mathbf{g}_m) + {\ell}_\textrm{ID}(\mathbf{C}_m) + {\ell}_\textrm{ID}(\mathbf{F}_m).
%+ {L}_\textrm{ID}(\mathbf{n}_m),
\label{eq:total id loss}
\end{equation}
% where $m$ is modality indicator, $\mathbf{C}_m= \{\mathbf{l}^{(i)}_m \}^{D}_{i=1}$, and $\mathbf{F}_m= \{\mathbf{p}^{(k)}_m \}^{P}_{k=1}$.
%where $\mathbf{C}_m=\{\mathbf{c}^{(i)}_m\}^{D}_{i=1}$, and $\mathbf{F}_m=\{\mathbf{f}^{(i)}_m\}^{P}_{i=1}$ % with modality indicator, $m$.

The second type is the triplet margin ranking loss~\cite{faghri2017vse++} for leaning joint image-text embedding space by ensuring that the negative sample is more dissimilar than the positive one by a margin $\alpha$.
% The second type is the triplet margin ranking loss~\cite{faghri2017vse++} for leaning joint image-text embedding space by ensuring that the negative sample is more dissimilar than the positive sample by a margin $\alpha$.
The learning objective for the ranking loss within a mini-batch of size $N$ is given by: 
% Let a mini-batch of image-text pairs be $\mathcal{B}=\{(e_i,T_i)\}_{i=1}^N$.
\begin{align}
\ell_\textrm{R}(\{\mathbf{e}_v^k,\mathbf{e}_t^k\}_{k=1}^N)&= \sum_{k=1}^{N}\Big([\alpha - s(\mathbf{e}^k_v,\mathbf{e}^k_t) + s(\mathbf{e}^k_v,\mathbf{e}^h_t)]_+ \nonumber \\
& +[\alpha - s(\mathbf{e}^k_t,\mathbf{e}^k_v) + s(\mathbf{e}^k_t,\mathbf{e}^h_v)]_+\Big),
% \mathcal{L}_\textrm{R}(\mathbf{E}_v,\mathbf{E}_t)&= [\alpha - s(\mathbf{E}^p_v,\mathbf{E}^p_t) + s(\mathbf{E}^p_v,\mathbf{E}^n_t)]_+ \nonumber \\
% & +[\alpha - s(\mathbf{E}^p_t,\mathbf{E}^p_v) + s(\mathbf{E}^p_t,\mathbf{E}^n_v)]_+,
\label{eq:ranking loss}
\end{align}
% \begin{align}
% \mathcal{L}_\textrm{R}(\mathbf{f}_v,\mathbf{f}_t)&= \textrm{max}(\alpha - s(\mathbf{f}^p_v,\mathbf{f}^p_t) + s(\mathbf{f}^p_v,\mathbf{f}^n_t),0) \nonumber \\
% & +\textrm{max}(\alpha - s(\mathbf{f}^p_t,\mathbf{f}^p_v) + s(\mathbf{f}^p_t,\mathbf{f}^n_v),0),
% \label{eq:ranking loss}
% \end{align}
\iffalse
where $[\cdot]_+\equiv \max(\cdot,0)$ and $s(\cdot,\cdot)$ is a cosine similarity between embeddings from two modalities, and $\mathbf{e}^h_t$ is the hardest negative for $\mathbf{e}^k_v$ in the mini-batch. 
Likewise, $\mathbf{e}^h_v$ is obtained by vice versa.
\fi
where $[\cdot]_+\equiv \max(\cdot,0)$ and $s(\cdot,\cdot)$ is a cosine similarity between embeddings from two modalities, $\mathbf{e}^h_t$ is the hardest negative for $\mathbf{e}^k_v$ in the mini-batch, and $\mathbf{e}^h_v$ is the hardest negative for $\mathbf{e}^k_t$.
The ranking loss applies to the set of embeddings except fine embeddings per modality $\mathcal{S}_m^\prime = \mathcal{S}_m~ \backslash~\mathbf{F}_{m}$:
\begin{eqnarray}
\mathcal{L}_\textrm{R}(\mathcal{S}_v^\prime,\mathcal{S}_t^\prime)={\ell}_\textrm{R}(\mathbf{g}_v,\mathbf{g}_t) + {\ell}_\textrm{R}(\mathbf{C}_v,\mathbf{C}_t).
%+ {L}_\textrm{R}(\mathbf{n}_v, \mathbf{n}_t).
\label{eq:total ranking loss}
\end{eqnarray}

However, when using this ranking loss, learning joint image-text embedding space with fine embeddings is challenging since a fine embedding of a person could be shared with other people, but the hardest negative for the fine embedding is selected based on the identity-level label.
To address this problem, we quantify the commonality of fine embeddings by calculating the entropy of the identity classification score distribution $\boldsymbol{p}$ in Eq.~\eqref{eq:id loss}.
The commonality is then reflected in the margin of the ranking loss. We call this ranking loss as commonality-based margin ranking (CMR) Loss.
The commonality $C$ of each fine embedding $\mathbf{f}_m\in\mathbf{F}_m$ is calculated by:
\begin{align}
    C(\mathbf{f}_m) &= -\sum_{i=1}^{c}p_{i}\log(p_{i}) /\log(c),
    % C(\mathbf{p}_m) &= H(\mathbf{p}_m) / \log(c),
    % C(\mathbf{p}_m) &= 1-H(\mathbf{p}_m) / \log(c),
    \label{eq:commonality}
\end{align}
where $p_{i}$ is the $i$-th identity classification score in Eq.~\eqref{eq:id loss}, and $c$ is the number of identities.
To estimate the commonality of fine embeddings, we normalize the entropy of the identity classification score distribution by dividing it by $\log(c)$. This ensures that the commonality value is between zero and one. 
% Then the commonality is quantified by the normalized entropy.
Lastly, the CMR loss is computed by:
\begin{align}
&\ell_\textrm{CMR}(\{\mathbf{f}_v^k,\mathbf{f}_t^k\}_{k=1}^N)\nonumber \\ 
&=\sum_{k=1}^N\Big([\alpha \cdot (1-C(\mathbf{f}_v^k)) - s(\mathbf{f}^k_v,\mathbf{f}^k_t) + s(\mathbf{f}^k_v,\mathbf{f}^h_t)]_+ \nonumber \\
& +[\alpha \cdot (1-C(\mathbf{f}_t^k)) - s(\mathbf{f}^k_t,\mathbf{f}^k_v) + s(\mathbf{f}^k_t,\mathbf{f}^h_v)]_+\Big).
% \mathcal{L}_\textrm{CMR}(\mathbf{p}_v, \mathbf{p}_t)&= \textrm{max}(\alpha \cdot C(\mathbf{p}_v) - s(\mathbf{p}^p_v,\mathbf{p}^p_t) + s(\mathbf{p}^p_v,\mathbf{p}^n_t),0) \nonumber \\
% & +\textrm{max}(\alpha \cdot C(\mathbf{p}_t) - s(\mathbf{p}^p_t,\mathbf{p}^p_v) + s(\mathbf{p}^p_t,\mathbf{p}^n_v),0).
\label{eq:cmr loss}
\end{align}

\begin{figure}[t!]
    \centering
    \includegraphics[width=0.6\textwidth]{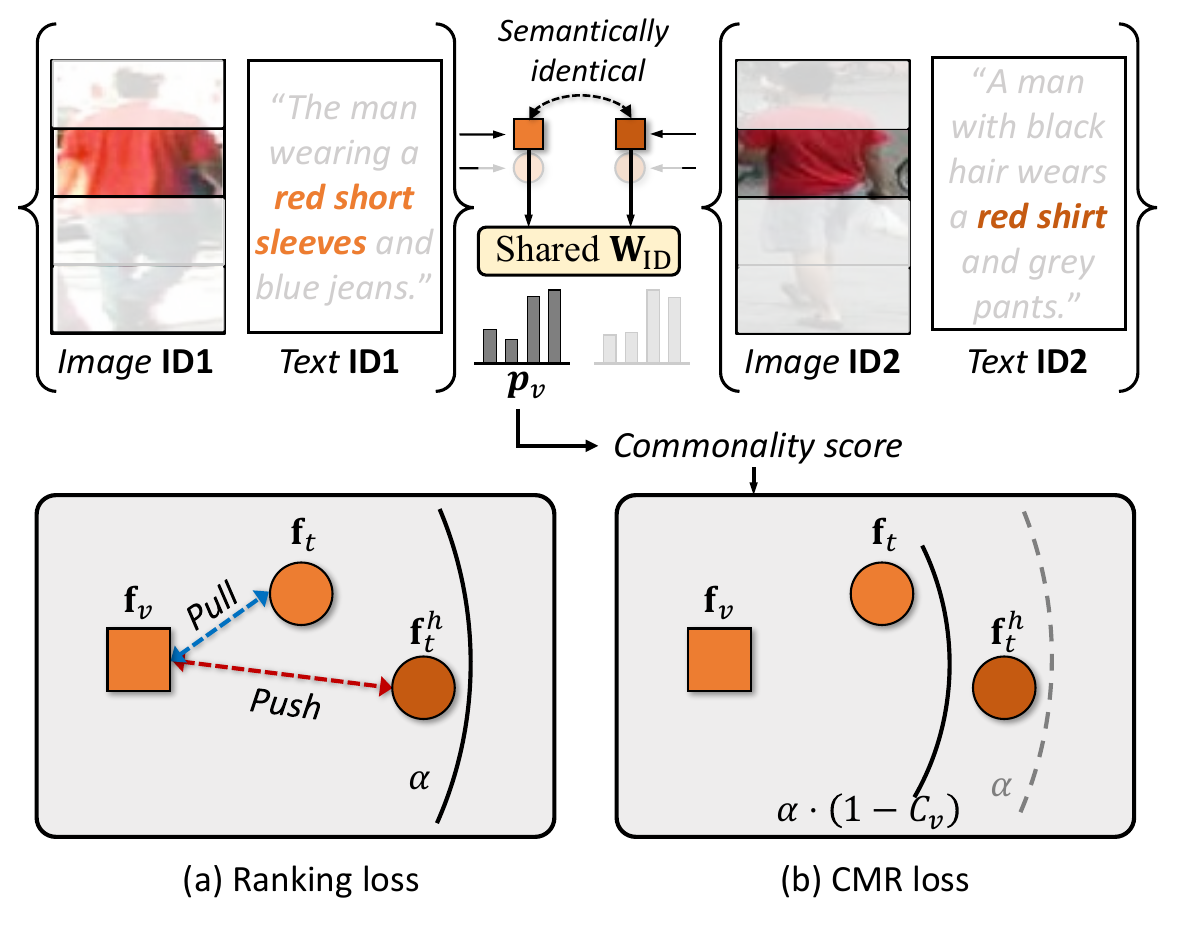}
    % \vspace{1mm} 
    \vspace{-3mm}
    \caption{A conceptual illustration of the Comanality-based Margin Ranking (CMR) loss function in Eq.~\eqref{eq:cmr loss}. The square and circle symbols denote the fine embeddings of image and text modalities, respectively.}
    \label{fig:Loss}
    \vspace{-3mm}
\end{figure}

% where $C_v$ and $C_t$ are commonality of each modality fine features $C(\mathbf{p}_v)$ and $C(\mathbf{p}_t)$, respectively.
The proposed CMR loss manages the margin with commonality to address the issue of semantically identical fine embeddings being learned as dissimilar due to identity-level labels.
The role of CMR loss is illustrated in Fig.~\ref{fig:Loss}.
Finally, our total loss is denoted by:
\begin{eqnarray}
% \mathcal{L}= \mathcal{L}_\textrm{ID}(\mathcal{S}) + \mathcal{L}_\textrm{R}(\mathcal{S}^\prime) + \mathcal{L}_\textrm{CMR}(\mathbf{F}).
\mathcal{L}= \mathcal{L}_\textrm{ID}(\mathcal{S}_v)+ \mathcal{L}_\textrm{ID}(\mathcal{S}_t) + \mathcal{L}_\textrm{R}(\mathcal{S}^\prime_v,\mathcal{S}^\prime_t) + \ell_\textrm{CMR}(\mathbf{F}_v,\mathbf{F}_t).
    \label{eq:total loss}
\end{eqnarray}

\subsection{Inference}\label{sec:inference}
During testing, the global, coarse, and fine embeddings of each modality input are fully exploited to calculate the similarity between the image-text pair.
Specifically, the similarity of the image-text pair, $S(\textit{I}, \textit{T})$, is defined as the sum of the similarities between the visual and the textual embeddings, which can be formulated as:
\begin{align}
S(\textit{I},\textit{T}) =& s(\mathbf{g}_v, \mathbf{g}_t) + \sum_{i=1}^{D}s\big(\mathbf{c}_v^{(i)}, \mathbf{c}_t^{(i)}\big) 
+ \sum_{j=1}^{P}s\big(\mathbf{f}_v^{(j)}, \mathbf{f}_t^{(j)}\big). %+ s(\mathbf{n}_v, \mathbf{n}_t).
\label{eq:sim}
\raisetag{5pt}
\end{align}
% where $s(\cdot,\cdot)$ denotes the cosine similarity.

Finally, given the text query, the images in the gallery are ranked according to similarity scores between the images and the text for inference.

%%inference using the sum of cosine sim between on each modality granularity.
% \vspace{1mm}\noindent\textbf{Commonality-Based Margin Ranking Loss.}
% \jicheol{
% The commonality of each fine feature is calculated by:
% \begin{eqnarray}
%     &H(\mathbf{f}) = -\sum_{id=1}^{c}p_{id}\log(p_{id})  \nonumber\\
%     &C(\mathbf{f}) = 1-H(\mathbf{f}) / \log(c)
%     \label{commonality}
% \end{eqnarray}
% where $p_{id}$ is the $id$-th identity score of \mathbf{f}

% The proposed loss mitigates the problem of semantically positive local features being dissimilar due to identity-level labels.
% }
% \subsection{Inference}~\label{sec:inference}
% \bstodo{Terminology (part embedding) be revised}
% During testing, the global and part embeddings of each modality input are fully exploited to calculate the similarity between the image-text pair. 
% The image-text pair similarity is defined as the sum of the similarity between the global embeddings of the image-text pair and the similarity between the part embeddings of it, which can be formulated as:
% \begin{equation}
%     S(\mathbf{T},\mathbf{V}) = \sum_{}^{} s(\mathcal{S}_t,\mathcal{S}_v).
%     \label{eq:sim}
% \end{equation}
% % where $s(\cdot,\cdot)$ denotes the cosine similarity.
% Finally, given the text query, the images in the gallery are ranked according to similarity scores between the images and the text for inference.
\vspace{-2mm}
\begin{table*}[!ht]
    \fontsize{6.5}{7.5}\selectfont
    \centering
    \caption{Performance comparison on the three datasets. \textbf{Bold} and \underline{underline} denote the best and the second best.}
    \resizebox{0.98\textwidth}{!}{%
    \begin{tabular}{l|cc|ccc|ccc|ccc}
    \toprule
    \multirow{2}{*}{Method}&\multicolumn{2}{c|}{Backbone}&\multicolumn{3}{c|}{CUHK-PEDES} &\multicolumn{3}{c|}{ICFG-PEDES}& \multicolumn{3}{c}{RSTPReid} \\
     &\multicolumn{1}{c}{Image} &\multicolumn{1}{c|}{Text} & \multicolumn{1}{c}{R@1} & \multicolumn{1}{c}{R@5} & \multicolumn{1}{c|}{R@10} & \multicolumn{1}{c}{R@1} & \multicolumn{1}{c}{R@5} & R@10 & \multicolumn{1}{c}{R@1} & \multicolumn{1}{c}{R@5} & \multicolumn{1}{c}{R@10} \\ \midrule
    % \multicolumn{10}{l}{\textit{\textbf{ResNet-50 and BERT}}} \\ [-0.3ex]\midrule
    GNA-RNN~\cite{li2017person} & RN50 & LSTM & 19.05 & - & 53.64 &-&-&-& -& -& -\\
    CMPM/C~\cite{zhang2018deep} & RN50 & LSTM  & 49.37 & 71.69 & 79.27 & 43.51& 65.44& 74.26&- &- &- \\
    PMA~\cite{jing2020pose}& RN50 & BERT  & 53.81 & 73.54 & 81.23 & - & - & - & -&- &- \\
    TIMAM~\cite{sarafianos2019adversarial} & RN101 & BERT  & 54.51 & 77.56 & 84.78 &- & -& -& -& -& -\\
    SCAN~\cite{lee2018stacked}& RN50 & BERT & 55.86 & 75.97 & 83.69 & 50.05& 69.65& 77.21& -&- &- \\
    ViTAA~\cite{wang2020vitaa}& RN50 & LSTM & 55.97 & 75.84 & 83.52 & 50.98& 68.79& 75.78& -& -&- \\
    NAFS~\cite{gao2021contextual}& RN50 & BERT & 59.94 & 79.86 & 86.70 & - & - & - & -&- &- \\
    DSSL~\cite{zhu2021dssl}& RN50 & BERT & 59.98 & 80.41 &87.56& - & - & - & 32.43 & 55.08 & 63.19 \\
    MGEL~\cite{wang2021text}& RN50 & LSTM & 60.27 & 80.01 & 86.74 & - & - & - & -&- &- \\
    SSAN~\cite{ding2021semantically}& RN50 & LSTM & 61.37 & 80.15 & 86.73 & 54.23 & 72.63 & 79.53& 43.50 & 67.80 & 77.15 \\
    LapsCore~\cite{wu2021lapscore}& RN50 & BERT & 63.40 & - & 87.80 & - & - & - & -&- &- \\
    %\hline \hline
    % IVT~\cite{ivt} & 64.00 & 82.72 & 88.95 & 56.04& 73.60& 80.22& 46.70& 70.00& 78.80 \\
    SRCF~\cite{suo2022simple} & RN50 & BERT & 64.04 & 82.99 & 88.81 & 57.18& 75.01& 81.49& - & -& -\\
    LGUR~\cite{lgur} & RN50 & BERT & 64.21 & 81.94 & 87.93 & 57.42 & 74.97 & 81.45 & -&- &- \\
    TIPCB~\cite{chen2022tipcb}& RN50 & BERT & 64.26 & \underline{83.19} & \underline{89.10} & - & - & - & -&- &- \\
    CAIBC~\cite{caibc}& RN50 & BERT & \underline{64.43} & 82.87 & 88.37 & - & - & - & \underline{47.35}& 69.55 & 79.00 \\
    BEAT~\cite{beat}& RN50 & BERT & 64.23 & 82.91 & 88.65 & \underline{57.62} & \underline{75.04} & \underline{81.53} & 47.30 & \underline{69.65} & \underline{79.20} \\
    %AXM-Net~\cite{axmnet}& RN50 & BERT & 64.44 & 80.52 & 86.77 & - & - & - & -& - & - \\
    % TextReID~\cite{han2021textreid} & 64.08 & 81.73 & 88.19 & - & - & - & & & \\  \hline
    % TIPCB~\cite{chen2022tipcb} & 64.26 & 83.19 & 89.10 & 54.96& 74.72& 81.89& & & \\  \hline
    % Ours (Global Only) & 65.09 & 82.60 & 89.10 & 56.95 & 74.79 & 81.36 & & & \\
    % Ours with AMT & 65.27 & 83.01 & 88.66 & 57.98 & 75.23 & 81.69 & 49.40& 72.50& 81.95 \\
     Ours & RN50 & BERT & \textbf{65.64} & \textbf{83.40} & \textbf{89.42} & \textbf{57.96} & \textbf{75.49} & \textbf{81.77} & \textbf{49.30} &\textbf{72.50} & \textbf{82.15} \\ \midrule
    % \multicolumn{10}{l}{\textit{\textbf{ViT-B/16 and BERT}}} \\ [-0.3ex]\midrule
    SAF~\cite{saf} & ViT-B/16 & BERT & 64.13 &82.62 &88.40 & - & - & - & - & - & -\\
    IVT~\cite{ivt} & ViT-B/16 & BERT & \underline{65.59} &\underline{83.11} &\underline{89.21} & \underline{56.04} &\underline{73.60} &\underline{80.22} & \underline{46.70} &\underline{70.00} &\underline{78.80}\\
    %Ours & \textbf{65.64} & \textbf{83.40} & \textbf{89.42} & \textbf{57.96} & \textbf{75.49} & \textbf{81.77} & \textbf{49.30} & \textbf{72.50} & \textbf{82.15} \\ \bottomrule
     Ours &  ViT-B/16 & BERT & \textbf{67.77} & \textbf{84.70} & \textbf{90.46} & \textbf{60.06}  & \textbf{76.37}  & \textbf{82.40} & \textbf{51.95}& \textbf{74.35}& \textbf{81.85} \\ \bottomrule    
    \end{tabular}
    }
    % \caption{Comparisons with state-of-the-art methods}
    %\caption{Performance of text-based person search methods on the three datasets. \textbf{Bold} and \underline{underline} denote the best and the second best.} %The performance which is not reported is denoted by -.}
    \label{tab:quan}
\end{table*}
\vspace{-9mm}

\section{Experiments}
\label{sec:Experiments}
In this section, we provide a detailed account of our experimental setup, evaluate our method and compare it with the state of the art on three benchmark datasets for text-based person search.

\subsection{Experimental Setup}
\label{subsec:experiment_setup}
\noindent \textbf{Datasets.} 
On three benchmark datasets, CUHK-PEDES \cite{li2017person}, ICFG-PEDES~\cite{ding2021semantically}, and RSTPReid~\cite{zhu2021dssl}, our method and previous methods are evaluated and compared.
In CUHK-PEDES which is collected from five existing person re-identification datasets~\cite{li2014deepreid, Market1501, xiao2016end, gray2007evaluating, li2013human}, there are 40,206 images from 13,003 person IDs and each image is approximately associated with the corresponding two annotated text descriptions.%, where each text description has 23 words at least.
We follow the data split of~\cite{li2017person} with 34,054 images from 11,003 person IDs and 68,126 text descriptions for training, 3,078 images from 1,000 person IDs and 6,158 text descriptions for validation, and 3,074 images from 1,000 person IDs and 6,156 text descriptions for testing.
The remaining two datasets are collected from MSMT17~\cite{wei2018person}.
ICFG-PEDES consists of 54,522 image-text pairs from 4,102 person IDs, which are split into 34,674 and 19,848 for training and testing, respectively.
RSTPReid contains 20,505 images of 4,101 person IDs from 15 cameras, where each person ID has 5 images, and each image is associated with the corresponding two annotated text descriptions.
We follow the data split of~\cite{zhu2021dssl} with 18,505 images from 3,701 person IDs and 37,010 text descriptions for training, 1,000 images from 200 person IDs and 2,000 text descriptions for validation, and 1,000 images from 200 person IDs and 2,000 text descriptions for testing, respectively.

\noindent \textbf{Evaluation Protocol.}
All experiments are evaluated by using the standard metric of rank at K (R@K=1,5,10) for fair comparisons to the previous work. 
% We set the K as 1, 5, and 10 for fair comparisons to the previous work.
Specifically, given a query text, all images are ranked according to their similarity scores.
The search is considered correct if at least one target image is placed within the first K ranks.

\noindent \textbf{Network Architecture.}
The training images are resized to 384$\times$128 and randomly flipped horizontally.
We adopt the ImageNet~\cite{Imagenet} pre-trained ResNet-50~\cite{resnet} and ViT-B/16~\cite{dosovitskiy2020image} as the backbone of the image while adopting the uncased BERT~\cite{devlin2018bert} for that of text.
% We adopt the ResNet-50~\cite{resnet} pre-trained on ImageNet~\cite{Imagenet} and uncased BERT~\cite{devlin2018bert} for the backbone of image modality and that of text modality, respectively. 
% \bstodo{To be revised:
% Each visual feature map extracted from the image backbone has 24$\times$ 8 $\times$ 2048 size.
% The text length and dimension of the textual embeddings are 64 and 2048, respectively.
% }

\noindent \textbf{Hyperparameters.}
During training, our model is optimized by Adam~\cite{Adamsolver} for 60 epochs. 
The mini-batch size is set to 32.
The initial learning rate is set to $5\mathrm{e}{-4}$ for our overall model, and it is decayed by a factor of 0.1 for 20, 40, 50, and 55 epochs.
The text backbone with BERT is not fine-tuned during training, while the image backbone is fine-tuned with a low learning rate by scaling 0.1 times.
In all experiments, we use the same hyperparameters for training.
The number of the tokens $D$ and the number of fine embeddings, $P$ are set to 4 and 4, respectively.

%\input{figures/fig_qual}
% \vspace{-3mm}

\subsection{Quantitative Results}
\label{subsec:experiment_quan}
Our method is compared to the existing state of the arts on CUHK-PEDES~\cite{li2017person}, ICFG-PEDES~\cite{ding2021semantically}, and RSTPReid~\cite{zhu2021dssl}.
Table \ref{tab:quan} shows the quantitative results of our method compared to state-of-the-art methods on these datasets. 
Ours using ResNet-50~\cite{resnet} as the image backbone achieves the best performance in terms of all evaluation metrics on all three datasets, with significant improvements over the previous state of the arts~\cite{caibc, chen2022tipcb, lgur, suo2022simple}.
Specifically, on the CUHK-PEDES dataset, our method outperforms the best previous method, CAIBC~\cite{caibc}, by 1.21\%p in terms of R@1.%, while ours also improves TIPCB~\cite{chen2022tipcb} by 0.22\%p and 0.32\%p in R@5 and R@10, respectively.
On ICFG-PEDES, our method surpasses BEAT~\cite{beat} by 0.34\%p in terms of R@1.%, and SRCF~\cite{suo2022simple} by 0.48\%p and 0.28\%p in R@5 and R@10, respectively.
Our method also improves the state-of-the-art scores on RSTPReid by a large margin, 1.95\%p, 2.15\%p, and 1.6\%p in terms of R@1, R@5, and R@10, respectively.

Moreover, our method using ViT-B/16~\cite{dosovitskiy2020image} as the image backbone achieves state-of-the-art performance with significant improvements over the previous SOTA~\cite{ivt}.
Specifically, ours outperforms IVT~\cite{ivt} by a large margin of 2.18\%p, 4.02\%p, and 5.25\%p in terms of R@1 on CUHK-PEDES, ICFG-PEDES, and RSTPReid, respectively.
% Notably, IVT exploits about 4M additional image-text pairs for pre-training the backbones, leading to obtaining better-generalized embeddings, while ours beats it without any image-text pairs for pre-training.
Notably, IVT exploits about 4M additional image-text pairs for pre-training the backbones, while ours beats it without any image-text pairs for pre-training.
% \iffalse
% Specifically, our method achieves 65.72\%, 83.46\%, and 89.31\% on R@1, R@5, and R@10, respectively, on CUHK-PEDES, outperforming the previous state-of-the-art method, SRCF~\cite{suo2022simple}, by 1.68\%, 0.47\%, and 0.5\%, respectively. 
% On ICFG-PEDES, our method achieves 58.15\%, 75.54\%, and 82.00\% on R@1, R@5, and R@10, respectively, which surpasses the previous state-of-the-art method by 0.97\%, 0.53\%, and 0.74\%. 
% Moreover, our method improves the state-of-the-art score on RSTPReid by a large margin, 4.65\%, 2.55\%, and 0.95\% in terms of R@1, R@5, and R@10, respectively.
% \fi

% These improvements demonstrate the superiority of our method which utilizes a modality-agnostic decoder framework to align the local features and considers the commonality of the local features (fine features), and thus it allows the embedding space to be more discriminative generalized.
The superior performance of our method can be attributed to its modality-sharing decoder framework, which enables aligning coarse embeddings from different modalities, and a novel ranking loss function that considers the commonality of fine embedding, allowing for learning a more discriminative embedding space.

%\addtocounter{footnote}{0}
%\footnotetext{We reproduced SRCF on the RSTPReid dataset through its official implementation available at \href{https://github.com/Suo-Wei/SRCF}{https://github.com/Suo-Wei/SRCF}.}

\begin{figure*}[t!]
    \centering
    \includegraphics[width=0.98\textwidth]{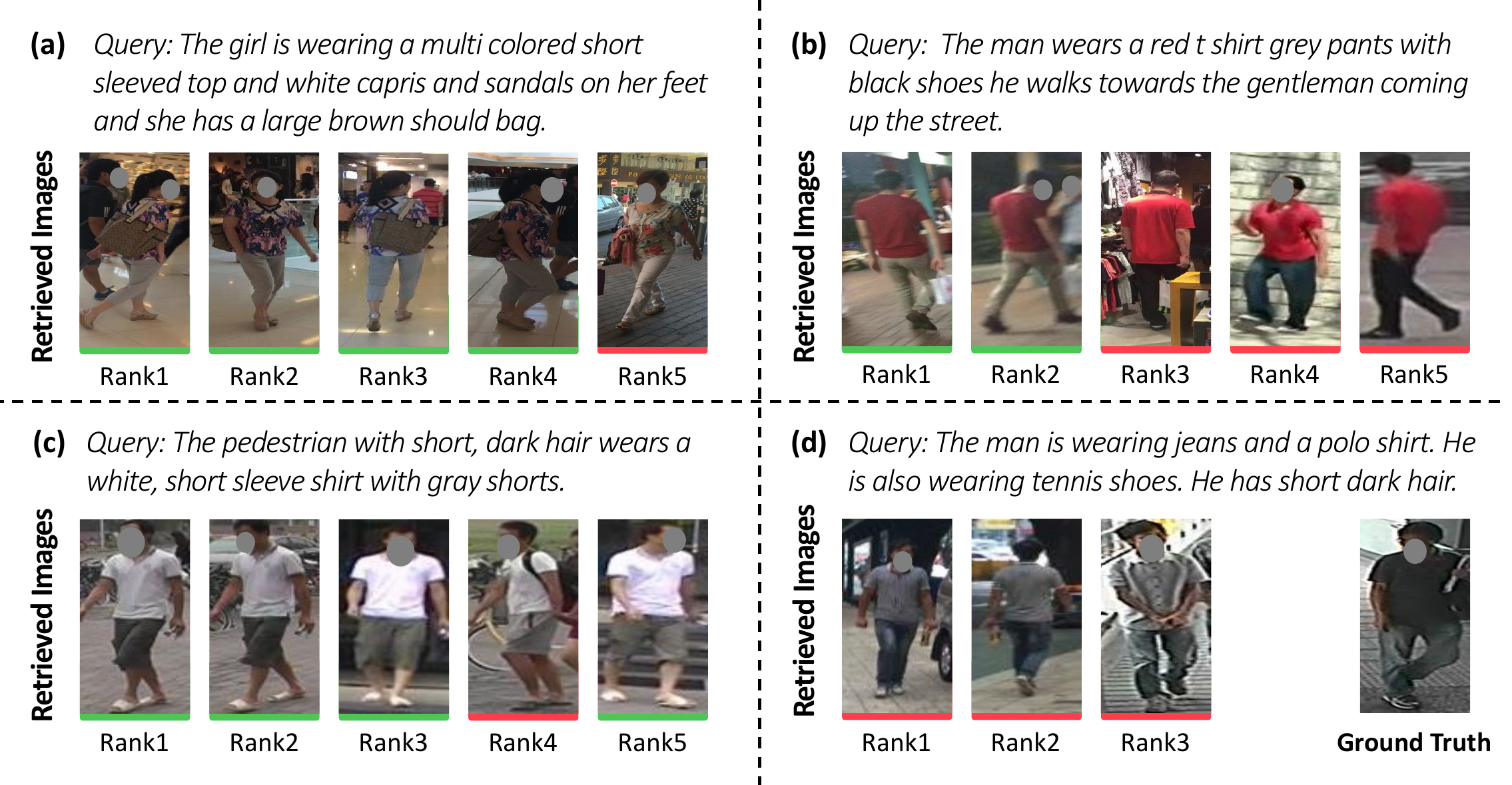}
    \caption{Qualitative results of our method on the CUHK-PEDES dataset. 
    Query texts and the retrieval results of our method for successful cases are presented, while the failure case of our method presents a query text, its ground truth, and the top 3 retrieval results.
    The true and false matches are colored green and red, respectively.
    }
    \label{fig:qualitative}
    %\vspace{-1mm}
\end{figure*}

\begin{figure*}[t!]
    \centering
    \includegraphics[width=0.98\textwidth]{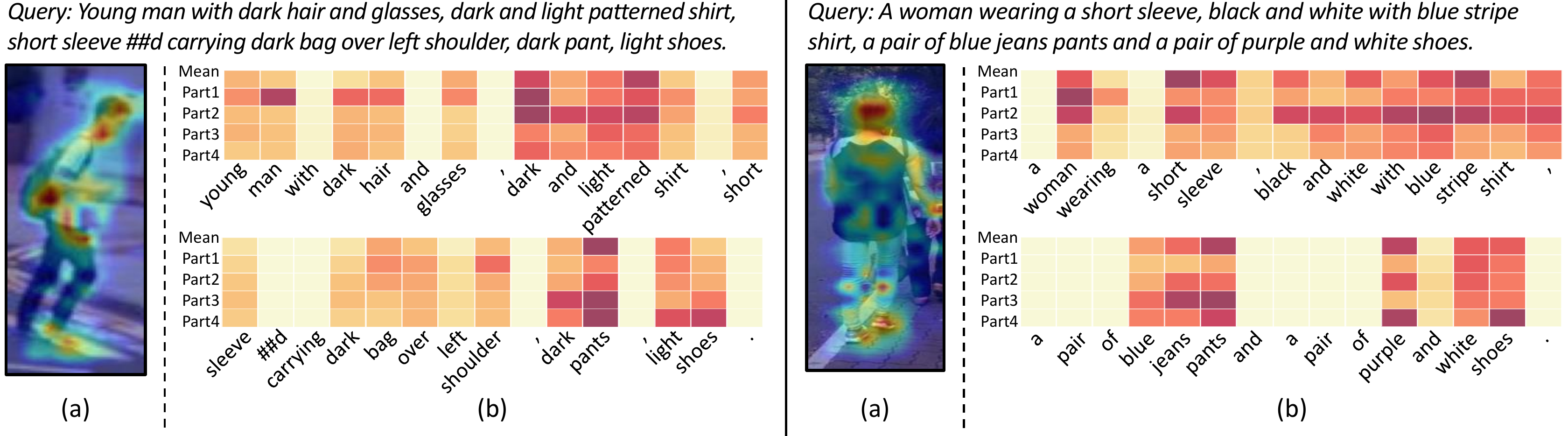}
    \caption{
    Visualization of a cross-attention map of (a) visual and (b) textual features from the decoder.
    }
    \label{fig:attention}
    \vspace{-4mm}
\end{figure*}

\subsection{Qualitative Results}
\label{subsec:experiment_qual}
%Qualitative results of our method on three datasets are presented in Fig.~\cref{fig:qualitative, fig:qual_icfg, fig:qual_rstp}, respectively.
Qualitative results of our method on CHHK-PEDES dataset are presented in  Fig.~\ref{fig:qualitative}.
Most of the presented results demonstrate that our method successfully retrieves the target images.
The individual examples presented in Fig.~\ref{fig:qualitative}(a,b,c), demonstrate that our method is able to distinguish hard negative samples, which partly match the given text descriptions.
Specifically, an image in Fig.~\ref{fig:qualitative}(a) that only differs from the query text description by ``\sftype{capris}'', and images in Fig.~\ref{fig:qualitative}(b) with different pants colors are retrieved at the end; in Fig.~\ref{fig:qualitative}(c), the image with a ``\sftype{backpack}'' not mentioned in the text is distinguished.
Failure case is also presented in Fig.~\ref{fig:qualitative}(d).
Note that false matches in the failure case are reasonable since the query text description is too ambiguous, which only describes a specific brand of cloth (\sftype{polo}) without any information on clothing colors.

To obtain a detailed insight into the aligning processes learned by our model, we visualize the cross-attentions from the decoder.
In Fig.~\ref{fig:attention}, the areas highlighted in red in both the image and text correspond to higher response scores. Fig.~\ref{fig:attention}(a) depicts the average of cross-attention weights from the decoder regarding the coarse embeddings of image modality.
The average of cross-attention weights $A_\textrm{avg}$ is shown to precisely cover human regions, reducing sensitivity to background clutter when extracting fine embeddings for the image modality, as denoted in Eq.~\eqref{spatial attended visual features}.
Fig.~\ref{fig:attention}(b) shows the cross-attentions from the decoder for text modality.
The first row exhibits the average cross-attention from the decoder with shared tokens and textual features, which covers a wide range of essential texts comprehensively.
The remaining rows represent cross-attentions from the decoder, which are computed using textual features and text tokens to extract fine embeddings for text modality.
Specifically, the first and second cross-attention primarily attend to text related to the upper body, such as ``\sftype{dark hair}'', ``\sftype{glasses}'', ``\sftype{short sleeve}'', ``\sftype{blue stripe shirt}'', and ``\sftype{dark and light patterned shirt}'', while the third and fourth cross-attention focus on text related to the lower body, such as ``\sftype{dark pants}'', ``\sftype{light shoes}'', ``\sftype{blue jeans pants}'', and ``\sftype{purple and white shoes}''.
These indicate that the fine embeddings for text modality are well aligned with the image-side fine embeddings, which are obtained by horizontally cutting from top to bottom.

% \addtocounter{footnote}{0}
% \footnotetext{We reproduced SRCF on the RSTPReid dataset through its official implementation available at \href{https://github.com/Suo-Wei/SRCF}{https://github.com/Suo-Wei/SRCF}.}

% \input{tables/tab_ablation}

% \input{tables/tab_decoder}

% \input{tables/tab_num_emb}

\subsection{Ablation Studies}
\label{subsec:experiment_ablation}
%\noindent \textbf{Effectiveness of Components.}
%To demonstrate the effectiveness of our proposed components, we conducted ablation studies on the CUHK-PEDES dataset~\cite{li2017person}.
To demonstrate the effectiveness of our proposed components, we conducted ablation studies on the CUHK-PEDES~\cite{li2017person}, ICFG-PEDES~\cite{ding2021semantically} and RSTPReid~\cite{zhu2021dssl} datasets with ResNet-50~\cite{resnet} as the image backbone.
We compared our method with different configurations and evaluated them on the retrieval performance.
We first build a fully reduced version of our method without all components removed denoted by ``Baseline".
It utilizes only global embeddings where the joint image-text embedding space is optimized by $\mathcal{L}_{\textrm{ID}}$ in Eq.~\eqref{eq:id loss} and $\mathcal{L}_{\textrm{R}}$ in Eq.~\eqref{eq:ranking loss}.
Based on it, the other variants are obtained by adding either coarse or fine embeddings.
%Among them, we define a baseline as the model that utilizes both global and coarse embeddings, where the number of tokens is set to 1.
%Except for the baseline, the number of tokens and the number of fine embeddings are set to 4 and 4, respectively.
We also examine the impact of $\mathcal{L}_{\textrm{CMR}}$ in Eq.~\eqref{eq:cmr loss} for learning a joint image-text embedding space with fine embeddings. 
In Table~\ref{tab:ablation_component}, we present the results of different ablation configurations.
When using only the global embeddings (the first row in Table~\ref{tab:ablation_component}), the performance is significantly low, indicating that global embeddings alone are not sufficient for accurate retrieval. 
Additionally, we analyze the effectiveness of our modality-sharing decoder and variations in the number of coarse embeddings.

\begin{table*}[!ht]
    \fontsize{7.5}{8.5}\selectfont
    \centering
    \caption{Ablation studies on the three public benchmark datasets. Coarse and Fine denote exploiting coarse embedding and fine embedding, respectively. CMR denotes that CMR loss is utilized to align visual and textual fine embeddings.}
    \resizebox{0.89\textwidth}{!}{   
    \begin{tabular}{l|ccc|ccc|ccc|ccc}
    \toprule
     \multirow{2}{*}{Method}&\multirow{2}{*}{Coarse}&\multirow{2}{*}{Fine}&\multirow{2}{*}{CMR} &\multicolumn{3}{c|}{CUHK-PEDES} &\multicolumn{3}{c|}{ICFG-PEDES}& \multicolumn{3}{c}{RSTPReid} \\
     &  & & & \multicolumn{1}{c}{R@1} & \multicolumn{1}{c}{R@5} & \multicolumn{1}{c|}{R@10} & \multicolumn{1}{c}{R@1} & \multicolumn{1}{c}{R@5} & \multicolumn{1}{c|}{R@10} &\multicolumn{1}{c}{R@1} & \multicolumn{1}{c}{R@5} & \multicolumn{1}{c}{R@10} \\ \midrule
    %Method & Coarse& Fine with CMR & R@1 & R@5 & R@10\\ \midrule
    Baseline & \xmark & \xmark & \xmark &59.76& 79.82& 86.45&54.39 & 72.69& 79.70& 44.20& 67.30&78.10 \\
    %Baseline & 1 & \xmark & \xmark& 62.52 & 81.35&88.09 \\ 
    & \checkmark &\xmark &\xmark &64.70& 83.07& 88.45& 56.92& 74.78& 81.32& 46.60&71.15&79.60 \\
    & \checkmark &\checkmark &\xmark &64.46& 83.07& 88.91& 57.03& 74.71& 81.34& 47.05&71.05&81.05 \\ \bottomrule
    %& \xmark& \checkmark& \xmark& 64.46 & 83.07 & 88.91 \\ \
    %& \xmark&4 &\checkmark & 65.09 & 83.27 & 88.65 \\
    %& \checkmark & \checkmark &\xmark & 64.96  & 82.37 & 88.87 \\ \midrule
    Ours & \checkmark  & \checkmark &\checkmark & \textbf{65.64} & \textbf{83.40} & \textbf{89.42} &\textbf{57.96} & \textbf{75.49} & \textbf{81.77} & \textbf{49.30} & \textbf{72.50} & \textbf{82.15} \\ \bottomrule
    \end{tabular}
    }
    \label{tab:ablation_component}
    \vspace{-2mm}
\end{table*}

\begin{table}[!t]
    %\vspace{1mm}
    \caption{Variant of ours with applying CMR loss at coarse and fine embeddings on the CUHK-PEDES dataset.}
    \vspace{-3mm}
    \centering
    \resizebox{0.60\textwidth}{!}{%   
    \begin{tabular}{c|l|ccc}
    \toprule
    \multicolumn{2}{l|}{Method}  & R@1 & R@5 & R@10\\ \midrule
    \multicolumn{2}{l|}{CMR applied to Coarse and Fine} & 64.99&  83.40 & 88.82 \\ \midrule
    %\multirow{2}{*}{(b)} &GAP & 62.20& 81.45 & 87.65\\
    % &GMP (ours) & 65.64 & 83.40 & 89.42 \\ 
    %&GAP+GMP & 65.12 & 82.70 & 88.81 \\ \midrule
    \multicolumn{2}{l|}{CMR applied to Fine only}& \textbf{65.64} & \textbf{83.40} & \textbf{89.42} \\ \bottomrule
    \end{tabular}
    }
    \label{tab:ablation_cmr}
    \vspace{-3mm}
\end{table}
\begin{table*}[!t]
   % \fontsize{5}{6}\selectfont
    \centering
    \caption{Performance of ours and its variant on the three datasets. \textbf{Bold} denotes the best performance.}
    \vspace{-3mm}
    \resizebox{0.90\textwidth}{!}{%
    \begin{tabular}{l|ccc|ccc|ccc}
    \toprule
    \multirow{2}{*}{Method}&\multicolumn{3}{c|}{CUHK-PEDES} &\multicolumn{3}{c|}{ICFG-PEDES}& \multicolumn{3}{c}{RSTPReid} \\
     & \multicolumn{1}{c}{R@1} & \multicolumn{1}{c}{R@5} & \multicolumn{1}{c|}{R@10} & \multicolumn{1}{c}{R@1} & \multicolumn{1}{c}{R@5} & R@10 & \multicolumn{1}{c}{R@1} & \multicolumn{1}{c}{R@5} & \multicolumn{1}{c}{R@10} \\ \midrule
    Separated Decoder & 64.75 & 82.70&88.60& 57.05&74.75&81.17&48.15&72.3&81.05 \\ 
     Ours & \textbf{65.64} & \textbf{83.40} & \textbf{89.42} & \textbf{57.96} & \textbf{75.49} & \textbf{81.77} & \textbf{49.30} & \textbf{72.50} & \textbf{82.15} \\ \bottomrule
    \end{tabular}
    }
    \label{tab:supp_decoder}
    \vspace{-5mm}
\end{table*}

\noindent \textbf{Effectiveness of Coarse Embeddings.}
%coarse embeddings, correspond local information 을 갖는, 사용했을때 significant improvements 확인 (compare 1st and 2nd row)
%fine embeddings, coarse 대비 뚜렷한 성능 향상을 보이진못함 (compare 2nd and 3rd)
%하지만 cmr통해 학습된 fine embeddings을 사용했을 경우에 뚜렷한 성능 향상을 확인 (compare 3rd and 4th)
By incorporating coarse embeddings that establish semantically aligned between modalities, substantial improvements were observed across all metrics.
This is clearly demonstrated through the differences between the first and second rows in Table~\ref{tab:ablation_component}.

\noindent \textbf{Appropriateness of CMR in Fine Embeddings.}
The second and third rows in Table~\ref{tab:ablation_component} reveal the marginal improvement of fine embeddings trained solely with the standard ranking loss $\mathcal{L}_{\textrm{R}}$.
However, as shown in the comparison between the third and fourth rows in Table~\ref{tab:ablation_component}, consistent and significant improvements were noted across all metrics for three benchmarks, when employing our CMR loss designed for the fine embedding learning. %$\mathcal{L}_{\textrm{CMR}}$
The reason for this is that the CMR loss deals with the possible existence of shared body parts among different individuals. 
% In other words, even when supervised with distinct identity labels, the CMR loss enables the training of fine embeddings representing semantically identical body parts to be close in the embedding space.
In other words, even when supervised with distinct identity labels, the CMR loss enables the fine embeddings representing semantically identical body parts to be close in the embedding space.
Moreover, to assess the appropriateness of the CMR loss, we compared its performance when applied to both coarse and fine embeddings, as well as when applied solely to fine embeddings.
The results are shown in the first and second rows in Table~\ref{tab:ablation_cmr}, respectively. Applying CMR solely to the fine embeddings that captured more specific information resulted in the most substantial performance improvement.
This is because the fine embeddings may contain more high common information, as the fine embeddings observe more specific areas. 
%As a result, fine embeddings may encounter problems that semantically identical fine embeddings are learned to be dissimilar under identity-level supervision.
As a result, fine embeddings are more likely to encounter the problem of semantically identical fine embeddings being learned as dissimilar under identity-level supervision.
This observation highlights the appropriateness of CMR in addressing the challenge of learning fine embeddings.

\noindent\textbf{Efficacy of Modality-sharing Decoder.} We demonstrate the efficacy of our modality-sharing decoder by comparing a variant that has individual decoders for each modality.
Table~\ref{tab:supp_decoder} shows that our modality-sharing decoder surpasses the variant with separate decoders for each modality on all evaluation metrics across three benchmarks, despite having fewer parameters. This suggests that the sharing decoder and tokens effectively align the semantics between modalities.

\noindent \textbf{Effectiveness of the Vision-Language Pre-trained Model.}
To enhance the text-based person search performance, we conduct experiments on the RSTPReid dataset using a CLIP pre-trained model~\cite{clip} as the backbone.
The CLIP pre-trained model is vision-language pre-trained model which has demonstrated remarkable performance in various vision and language tasks.
By leveraging the joint representation of images and texts from the pre-trained CLIP, we effectively bridge the gap between visual and textual modalities.
The experiment results are shown in Table~\ref{tab:clip}.
Additionally, we compared our method to IRRA~\cite{irra}, a state-of-the-art method that also uses the CLIP pre-trained model as its backbone.
As a result, we achieved significant performance improvements by leveraging the vision-language pre-trained model as the backbone.
Moreover, we outperformed IRRA in the R@1 and R@5 metrics.

\vspace{-4mm}
\begin{table*}[!ht]
    \fontsize{6.5}{7.5}\selectfont
    \centering
    \caption{Ablation study on different backbone models and performance comparison on RSTPReid dataset with IRRA.}
    \vspace{-2mm}
    \resizebox{0.75\textwidth}{!}{%
    \begin{tabular}{l|cc|ccc}
    \toprule
    \multirow{2}{*}{Method}&\multicolumn{2}{c|}{Backbone}&\multicolumn{3}{c}{RSTPReid}  \\
     &\multicolumn{1}{c}{Image} &\multicolumn{1}{c|}{Text} & \multicolumn{1}{c}{R@1} & \multicolumn{1}{c}{R@5} & \multicolumn{1}{c}{R@10}  \\ \midrule
    % \multicolumn{10}{l}{\textit{\textbf{ResNet-50 and BERT}}} \\ [-0.3ex]\midrule
    Ours & RN50 & LSTM & 49.30 & 72.50 & 82.15\\ \midrule
    Ours & ViT-B/16 &  BERT & 51.95 & 74.35 & 81.85  \\ \midrule
    % \multicolumn{10}{l}{\textit{\textbf{ViT-B/16 and BERT}}} \\ [-0.3ex]\midrule
    IRRA~\cite{irra} & CLIP-ViT-B/16 & CLIP-Xformer & \underline{60.20} &\underline{81.30} &\textbf{88.20} \\
    Ours &  CLIP-ViT-B/16 & CLIP-Xformer & \textbf{60.90} & \textbf{81.65} & \underline{88.10}  \\ \bottomrule    
    \end{tabular}
    }
    \label{tab:clip}
    \vspace{-9mm}
\end{table*}
\section{Conclusion and Discussion}
In this paper, we have proposed an encoder-decoder architecture based on multi-head attention and a novel loss function.
The encoder-decoder model enables reducing the modality gap by extracting multiple person embeddings semantically aligned between the image and text modalities with shared tokens and decoder.
% 
% a set of latent local features and enables aligning them without dedicated local supervision by sharing tokens and the decoder across modalities.
Also, the new loss function allows our model to learn to fine-grained information appropriately with only ID-level supervision. This enhancement contributes to the model's ability to capture subtle differences effectively.
We demonstrated by extensive experiments that our method allows for achieving the state of the art.

Our method showed promising results in the current text-based person search task, but there are a couple of limitations that need to be discussed.
Firstly, current benchmarks in text-based person search predominantly rely on images that are tightly cropped around individuals. This approach, however, does not accurately reflect real-world scenarios where cameras used for person search typically capture wide scenes, not just isolated objects. Integrating person detection with the proposed model represents a promising avenue for future development in this field.
Secondly, the current method uniformly divides images in the vertical direction to extract visual fine embeddings, presupposing pedestrians are always in an upright posture.
Incorporating a learnable part-level decomposition module into the proposed model could lead to a more robust system, capable of effectively handling such challenges. This enhancement represents another promising direction for research.

\clearpage  % TODO REVIEW/FINAL: This \clearpage needs to be removed from both review and camera-ready versions.

\bibliographystyle{splncs04}
%\bibliography{main}
\bibliography{cvlab_kwak}

% ---------------------------------------------------------------
% Hyperref package

% It is strongly recommended to use hyperref, especially for the review version.
% Please disable hyperref *only* if you encounter grave issues.
% hyperref with option pagebackref eases the reviewers' job, but should be disabled for the final version.
%
% If you comment hyperref and then uncomment it, you should delete
% main.aux before re-running LaTeX.
% (Or just hit 'q' on the first LaTeX run, let it finish, and you
%  should be clear).

% TODO FINAL: Comment out the following line for the camera-ready version

%%%%%%%%%%%%%%%%%%%%%
\newcounter{alphasect}
\def\alphainsection{0}

\let\oldsection=\section
\def\section{%
  \ifnum\alphainsection=1%
    \addtocounter{alphasect}{1}
  \fi%
\oldsection}%

\renewcommand\thesection{%
  \ifnum\alphainsection=1% 
    \Alph{alphasect}
  \else%
    \arabic{section}
  \fi%
}%

\newenvironment{alphasection}{%
  \ifnum\alphainsection=1%
    \errhelp={Let other blocks end at the beginning of the next block.}
    \errmessage{Nested Alpha section not allowed}
  \fi%
  \setcounter{alphasect}{0}
  \def\alphainsection{1}
}{%
  \setcounter{alphasect}{0}
  \def\alphainsection{0}
}%
%%%%%%%%%%%%%%%%%%%%%%

\title{Improving Text-based Person Search via\\Part-level Cross-modal Correspondence\\\emph{--- Supplementary Materials ---}}


%\author{First Author\inst{1}\orcidlink{0000-1111-2222-3333} \and
%Second Author\inst{2,3}\orcidlink{1111-2222-3333-4444} \and
%Third Author\inst{3}\orcidlink{2222--3333-4444-5555}}

%\authorrunning{F.~Author et al.}

%\institute{Princeton University, Princeton NJ 08544, USA \and
%Springer Heidelberg, Tiergartenstr.~17, 69121 Heidelberg, Germany
%ABC Institute, Rupert-Karls-University Heidelberg, Heidelberg, %Germany\\

\author{Jicheol Park \orcidlink{0009-0004-7899-6802} \and
Boseung Jeong \orcidlink{0000-0001-9382-3396} \and
    Dongwon Kim \orcidlink{0000-0003-1147-5274} \and
    Suha Kwak \orcidlink{0000-0002-4567-9091}}

\authorrunning{J.~Park et al.}

\institute{ Pohang University of Science and Technology (POSTECH), South Korea\\
\email{\{jicheol, boseung01, kdwon, suha.kwak\}@posetech.ac.kr
}}
\maketitle

\begin{alphasection}
\renewcommand{\thefigure}{\Alph{figure}}

This supplementary material commences with an outline of the notations employed in the main paper. 
Thereafter, we offer an ablation study on varying the number of coarse embeddings. 
Lastly, we furnish more qualitative results for all three benchmark datasets.

\vspace{-3mm}\section{Notation and More Ablation Study}
\vspace{2mm}\noindent\textbf{Notation.}
Table~\ref{tab:notation table} shows outlines of the notations used in the main paper and their respective definitions.
% \section{Notation}
%\setcounter{table}{0}
% \renewcommand{\thetable}{A.\arabic{table}}
%\counterwithin{table}{section}
\vspace{-4mm}
\begin{table}[!ht]
    \caption{Notation repository: A comprehensive reference table containing notations and their corresponding explanations.}
    \centering
    \resizebox{0.81\textwidth}{!}{%
    \begin{tabular}{cc|c}
    \toprule
    \multicolumn{2}{c|}{Notation}&\multirow{2}{*}{Description}\\ 
    Image&Text& \\ \midrule 
    $\mathbf{V}$& $\mathbf{T}$ & \multicolumn{1}{l}{Output features of each backbone}  \\ \midrule
    \multirow{2}{*}{$\mathbf{g}_v$}& \multirow{2}{*}{$\mathbf{g}_t$} & \multicolumn{1}{l}{Global embeddings} \\
    && \multicolumn{1}{l}{(global max pooling of $\mathbf{V}$ and $\mathbf{T}$)}\\ \midrule
    $\bar{\mathbf{V}}$& - & \multicolumn{1}{l}{Flattened features of $\mathbf{V}$ }  \\ \midrule 
    $\mathbf{E}^{pos}_v$& $\mathbf{E}^{pos}_t$ & \multicolumn{1}{l}{Learnable position embedding}  \\ \midrule
    $\mathbf{M}_v$& $\mathbf{M}_t$ & \multicolumn{1}{l}{Position-encoded features}  \\ \midrule 
    \multirow{2}{*}{$\tilde{\mathbf{M}}_v$}& \multirow{2}{*}{$\tilde{\mathbf{M}}_t$} & \multicolumn{1}{l}{Output features of modality-specific encoder }  \\ 
    &&\multicolumn{1}{l}{(self-attended features)}\\\midrule 
    \multicolumn{2}{c|}{$\mathbf{D}$} & \multicolumn{1}{l}{Shared a set of learnable tokens }  \\ \midrule
    \multirow{2}{*}{$\mathbf{C}_v$}& \multirow{2}{*}{$\mathbf{C}_t$} & \multicolumn{1}{l}{Coarse embeddings}\\
    &&\multicolumn{1}{l}{(output features of modality-sharing decoder)}  \\ \midrule 
    $A_\textrm{avg}$& - & \multicolumn{1}{l}{Average of cross-attention weights between $\tilde{\mathbf{M}}_v$ and $\mathbf{D}$}  \\ \midrule
    % $A_\textrm{avg}$& - & \multicolumn{1}{l}{Average cross attention between $\tilde{\mathbf{M}}_v$ and $\mathbf{D}$}  \\ \midrule
    $\ddot{\mathbf{M}}_v$&  - & \multicolumn{1}{l}{Foreground visual features of $\tilde{\mathbf{M}}_v$ with $A_\textrm{avg}$}\\ \midrule 
    -& $\mathbf{D}_t$ & \multicolumn{1}{l}{Additional learnable tokens for text modality}  \\ \midrule
    $\mathbf{F}_v$& $\mathbf{F}_t$ & \multicolumn{1}{l}{Fine embeddings}   \\ \midrule
    {$\mathcal{S}_v$}& {$\mathcal{S}_t$} & \multicolumn{1}{l}{A set of embeddings including global $\mathbf{g}$, coarse $\mathbf{C}$, and fine $\mathbf{F}$.} \\ \bottomrule
    %$\mathbf{n}_v$& $\mathbf{n}_t$ & \multicolumn{1}{l}{Non-local embeddings}  \\ \bottomrule
    \end{tabular}
    }
    \label{tab:notation table}
    \vspace{-4mm}
\end{table}

\noindent\textbf{Variation on the number of Coarse Embeddings.}
\vspace{4mm}%The Tab. ~\ref{tab:supp_num} shows the performance of variants with different number of coarse and fined embeddings
Table ~\ref{tab:supp_num} shows the results of an ablation study that investigated the impact of different numbers of coarse embeddings on three benchmark datasets. Irrespective of the number of coarse embeddings employed, our proposed method surpasses previous studies on all three datasets, as demonstrated in Table ~\ref{tab:supp_num}. Ultimately, the optimal number of coarse embeddings is determined to be 4, as per our observation.
\begin{table*}[!t]
    \fontsize{5}{5.5}\selectfont
    \centering
    \caption{Performance of variants with different numbers of coarse embeddings on the three datasets. \textbf{Bold} denotes the best performance.}
    \resizebox{0.89\textwidth}{!}{%
    \begin{tabular}{l|c|ccc|ccc|ccc}
    \toprule
    \multirow{2}{*}{Method}& \multirow{2}{*}{Coarse} &\multicolumn{3}{c|}{CUHK-PEDES} &\multicolumn{3}{c|}{ICFG-PEDES}& \multicolumn{3}{c}{RSTPReid} \\
     & &  \multicolumn{1}{c}{R@1} & \multicolumn{1}{c}{R@5} & \multicolumn{1}{c|}{R@10} & \multicolumn{1}{c}{R@1} & \multicolumn{1}{c}{R@5} & R@10 & \multicolumn{1}{c}{R@1} & \multicolumn{1}{c}{R@5} & \multicolumn{1}{c}{R@10} \\ \midrule
    & 2&  65.20 & 82.54 & 89.13 & 57.69 & 74.59 & 81.32 & 47.60 & 70.30 & 81.95\\
     Ours & 4& \textbf{65.64} & \textbf{83.40} & \textbf{89.42} & \textbf{57.96} & \textbf{75.49} & \textbf{81.77} & \textbf{49.30} & \textbf{72.50} & \textbf{82.15} \\     
    & 6&  65.02 & 82.98 & 88.98 & 57.49 & 75.14 & 81.57 & 49.05 & 71.55 & 80.70 \\\bottomrule
    \end{tabular}
    }
    \label{tab:supp_num}
\end{table*}

\vspace{3mm}\section{More Qualitative Results}
We furnish more qualitative results of our method on three datasets that are presented in Fig.~\ref{fig:qual_cuhk}, Fig.~\ref{fig:qual_icfg}, and Fig.~\ref{fig:qual_rstp}, respectively.
Most of the presented results illustrate that our method successfully retrieves the target images.
Specifically, Fig.~\ref{fig:qual_cuhk}(a,b) presents examples that demonstrate the capability of our model to retrieve target images successfully even when there is significant background clutter.
Moreover, our model successfully recognizes fine-grained characteristics as shown in Fig.~\ref{fig:qual_cuhk}(a,d), Fig.~\ref{fig:qual_icfg}(d,h), and Fig.~\ref{fig:qual_rstp}(g,i), such as \sftype{wearing sun glasses}, \sftype{smoking a cigarette}, \sftype{black backpack with white dots design}, \sftype{white shopping bags in her hand}, \sftype{loose grey pants}, and \sftype{her hand in her pocket}.
Furthermore, the failure cases shown in Fig.~\ref{fig:qual_cuhk}-~\ref{fig:qual_rstp} are understandable because the falsely matched images exhibit subtle differences from the target images. For example, the presence of absence of a bag (Fig.~\ref{fig:qual_cuhk}(e) and Fig.~\ref{fig:qual_icfg}(j)) and variations in clothing color (Fig.~\ref{fig:qual_icfg}(j) and Fig.~\ref{fig:qual_rstp}(j)).

\begin{figure*}[!hb]
    \vspace{-10mm}
    \centering
    \includegraphics[width=0.95\linewidth]{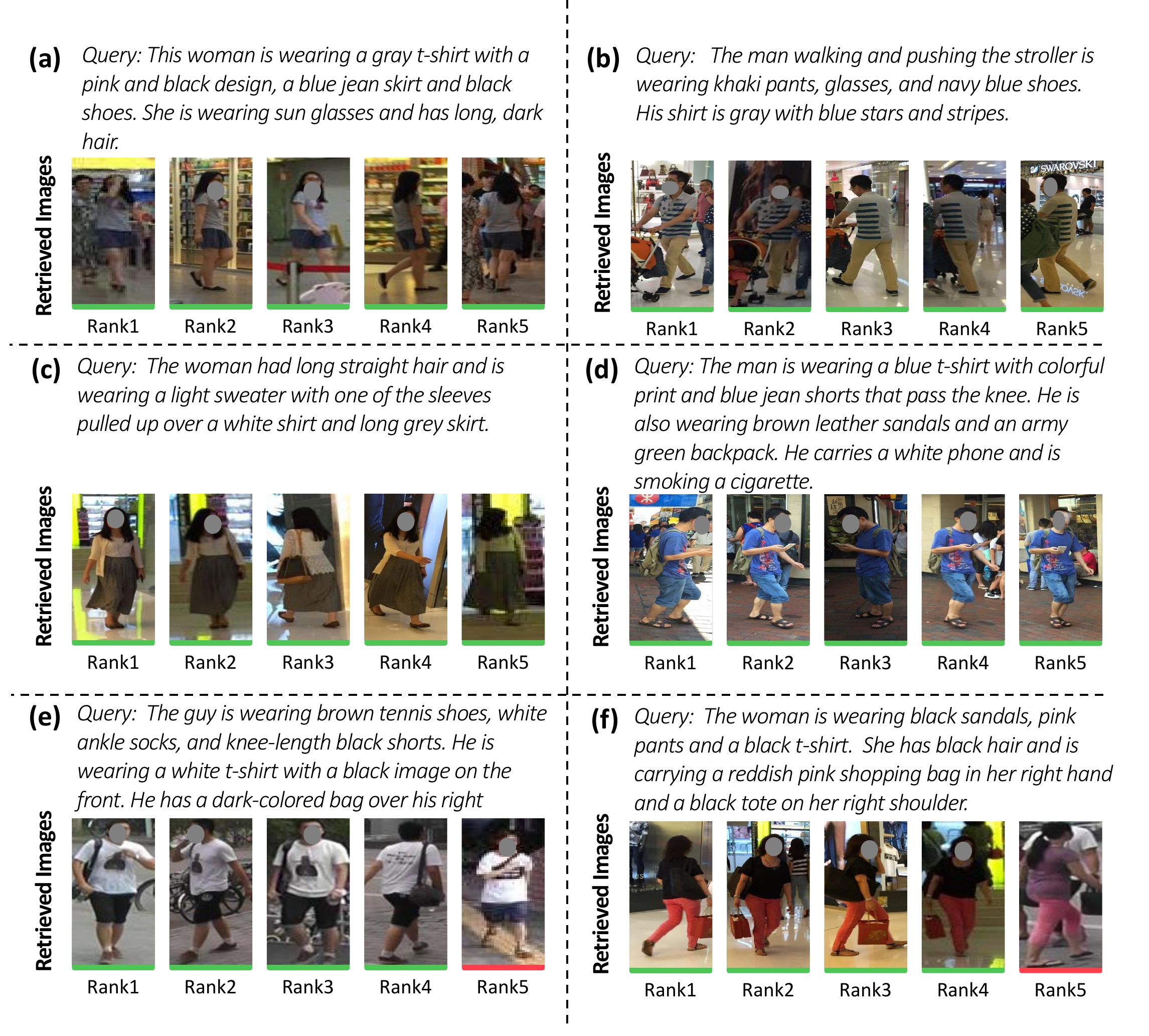}
    % \vspace{-5mm}
    \caption{
    Qualitative results of our method on the CUHK-PEDES dataset. 
    Query texts and the top-5 retrieval results of our method are presented, 
    The true and false matches are colored green and red, respectively.
    }
    \vspace{-5mm}
    \label{fig:qual_cuhk}
\end{figure*}

\renewcommand{\thefigure}{\Alph{figure}}

%\newpage
\begin{figure*}[t!]
    \centering
    \includegraphics[width=0.95\textwidth]{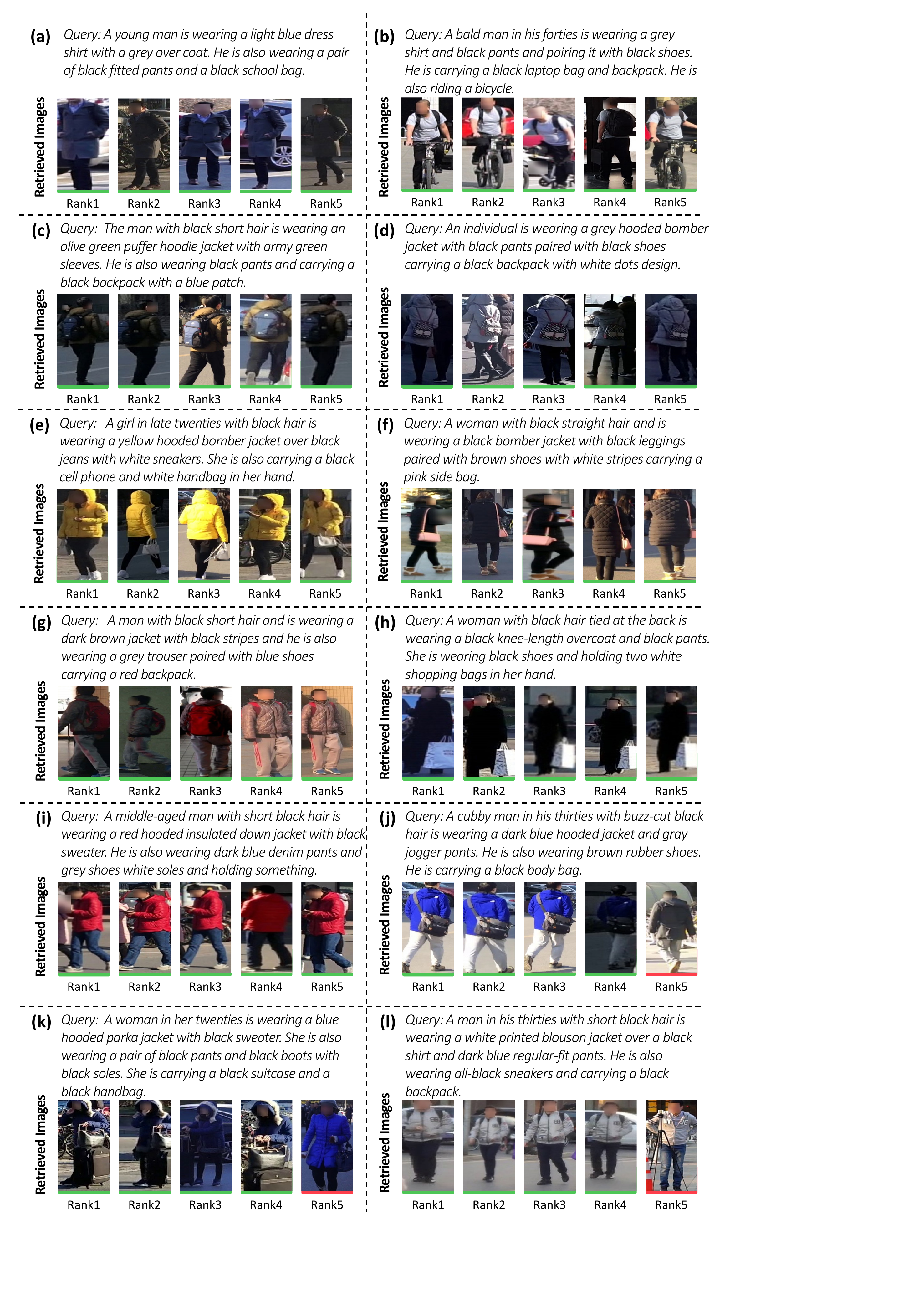}
    \vspace{-2mm}
    \caption{
    Qualitative results of our method on the ICFG-PEDES dataset. 
    Query texts and the top-5 retrieval results of our method are presented, 
    The true and false matches are colored green and red, respectively.
    }
    \label{fig:qual_icfg}
\end{figure*}

% \newpage
\begin{figure*}[t!]
    \centering
    \includegraphics[width=0.95\textwidth]{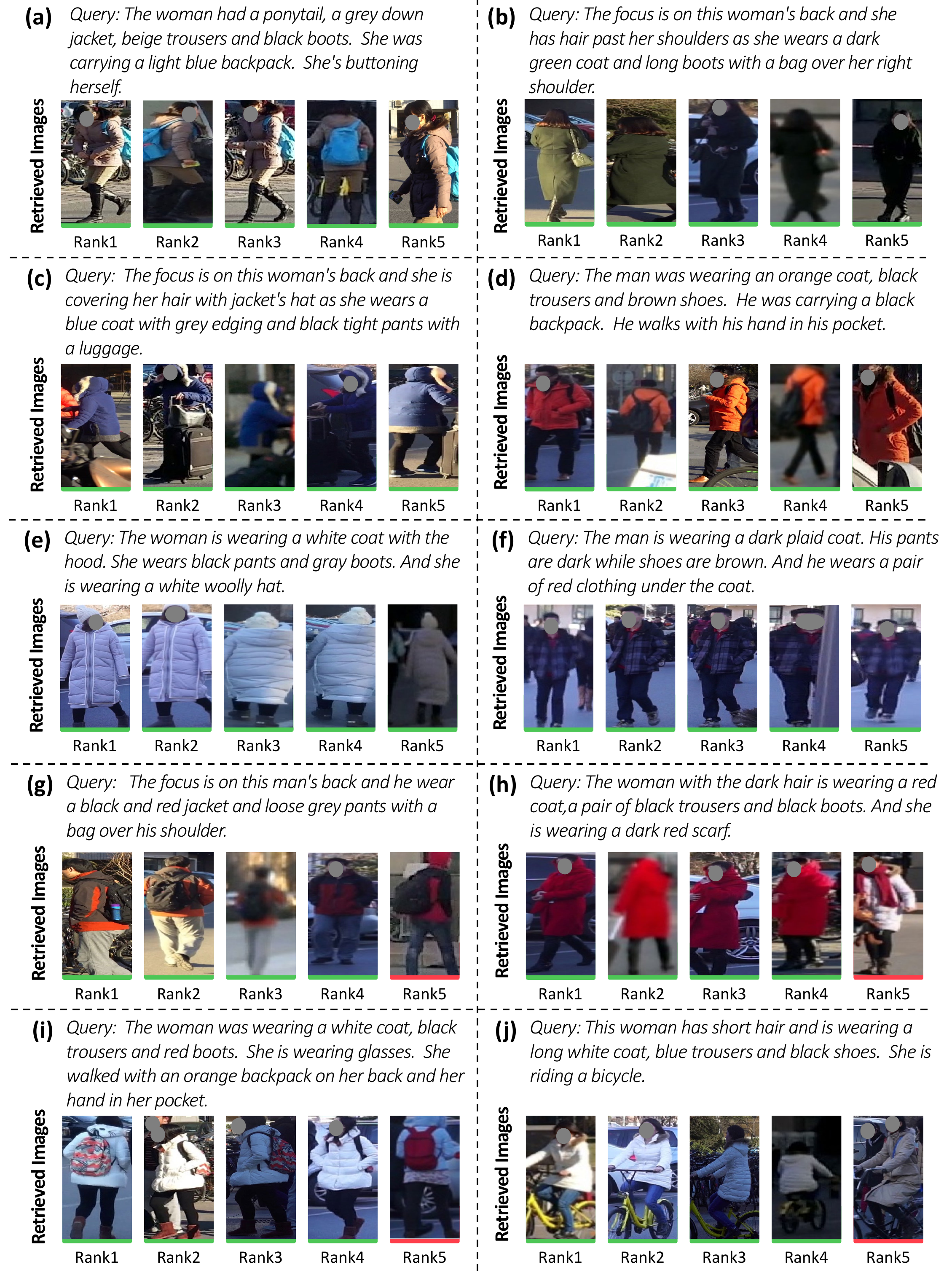}
    \vspace{-2mm}
    \caption{
    Qualitative results of our method on the RSTPReid dataset. 
    Query texts and the top-5 retrieval results of our method are presented, 
    The true and false matches are colored green and red, respectively.
    }
    \label{fig:qual_rstp}
\end{figure*}

\end{alphasection}

%\clearpage  % TODO REVIEW/FINAL: This \clearpage needs to be removed from both review and camera-ready versions.

%\bibliographystyle{splncs04}
%\bibliography{main}
%\bibliography{cvlab_kwak}

\end{document}